\documentclass[11pt]{article}
\usepackage{acl}
\usepackage{graphicx}
\usepackage{hyperref}
\usepackage{booktabs}
\usepackage{multirow}
\usepackage{tabularx} 
\usepackage{array} 
\usepackage{siunitx}
\usepackage[most]{tcolorbox}
\usepackage{amsmath}
\usepackage{longtable}
\usepackage{subcaption}
\usetikzlibrary{shapes,arrows,positioning,fit,backgrounds}

\title{Decoupling Scores and Text: The Politeness Principle in Peer Review}

\author{Yingxuan Wen\\
    Harbin Intitute Of Technology \\
  \texttt{2023112752@stu.hit.edu.cn} \\}

\begin{document}
\maketitle

\begin{abstract}
Authors often struggle to interpret peer review feedback, deriving false hope from polite comments or feeling confused by specific low scores. To investigate this, we construct a dataset of over 30,000 ICLR 2021–2025 submissions and compare acceptance prediction performance using numerical scores versus text reviews. Our experiments reveal a significant performance gap: score-based models achieve 91\% accuracy, while text-based models reach only 81\% even with large language models, indicating that textual information is considerably less reliable. To explain this phenomenon, we first analyze the 9\% of samples that score-based models fail to predict, finding their score distributions exhibit high kurtosis and negative skewness, which suggests that individual low scores play a decisive role in rejection even when the average score falls near the borderline. We then examine why text-based accuracy significantly lags behind scores from a review sentiment perspective, revealing the prevalence of the Politeness Principle: reviews of rejected papers still contain more positive than negative sentiment words, masking the true rejection signal and making it difficult for authors to judge outcomes from text alone.
\footnote{Our code and data are publicly available at \url{https://github.com/amanda8170/Review-Analysis/tree/main}}.
\end{abstract}

% \noindent 
% \textbf{Keywords:}
% Peer Review Analysis, Acceptance Prediction, Large Language Models, ICLR.

\section{Introduction}

Peer review serves as the gatekeeping mechanism for scientific progress. Ideally, the quantitative ratings and qualitative text in a review should be consistent. However, in practice, authors often face difficulties in interpreting peer review feedback. They may derive false hope from polite comments or feel confused by specific low scores despite generally positive remarks. The transparency of platforms such as OpenReview offers a unique opportunity to scrutinize this consistency and help authors objectively understand decision signals.

The digitalization of this process has spurred extensive computational research utilizing datasets such as PeerRead \cite{kang-etal-2018-dataset} to model acceptance prediction. Previous works have predominantly approached this challenge as a text classification task, employing techniques ranging from aspect-based sentiment analysis \cite{10.1145/3209978.3210056, MinghuiMeng2023AspectbasedSA} to advanced deep learning architectures \cite{10.1145/3383583.3398541}. Recent studies have also explored the dynamics of the rebuttal phase \cite{kargaran2025insights}. However, few studies have ally quantified the reliability of text-based prediction compared to score-based prediction in the era of large-scale open reviewing.
%%%%%%%%%%%%%%%%%%%%%%%%%%%%%%%%%%%%
\begin{figure*}[!ht]
    \centering
    \includegraphics[width=0.85\linewidth]{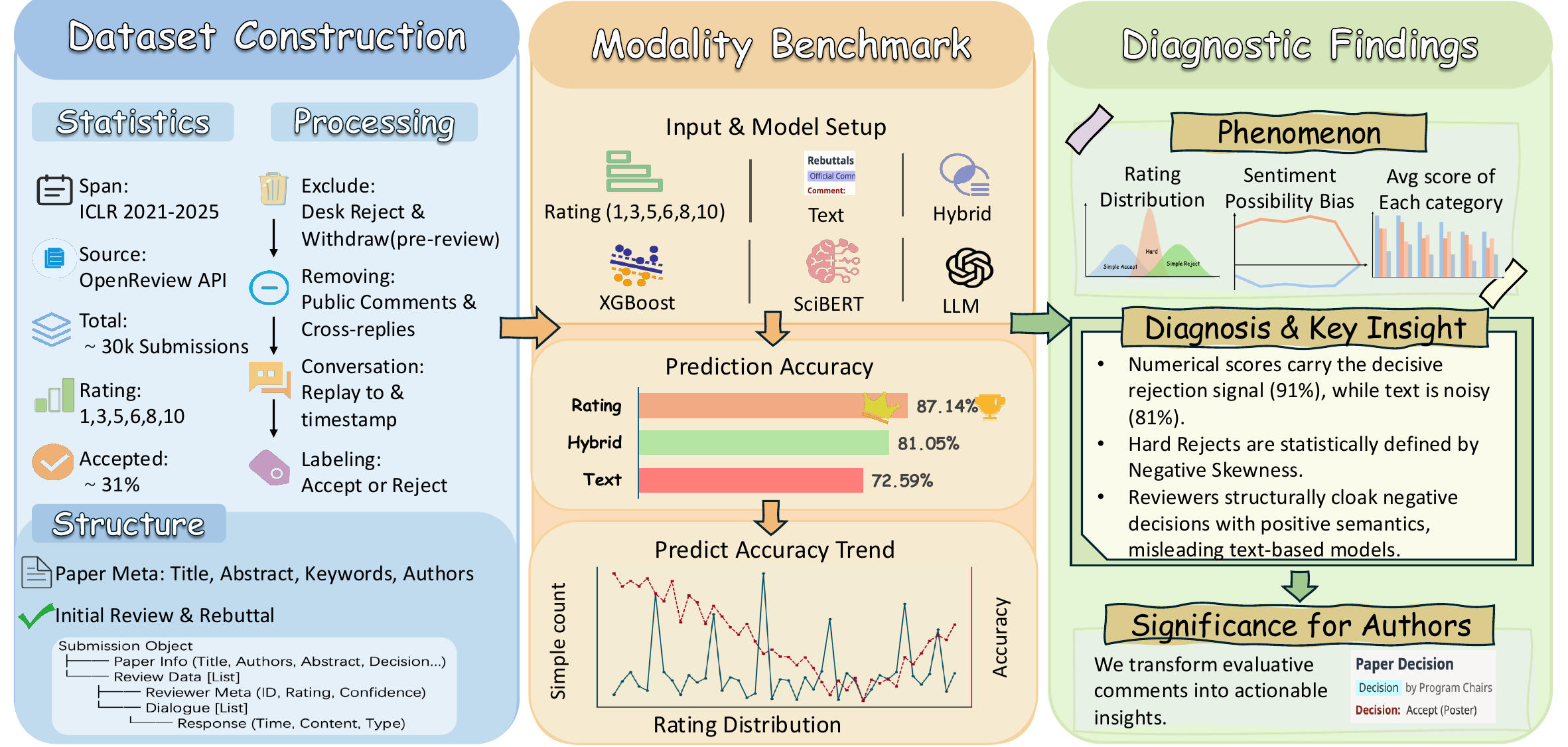}
    \caption{Overview of the research framework. It integrates large-scale dataset construction, multi-modality acceptance prediction benchmarking (Score vs. Text), and a diagnostic analysis of hard samples to quantify the impact of the Politeness Principle on peer review.}
    \label{fig:framework}
\end{figure*}
%%%%%%%%%%%%%%%%%%%%%%%%%%%%%%%%%%%%

To address this gap, we constructed a comprehensive dataset from ICLR 2021-2025, covering over 30,000 submissions. We utilized acceptance prediction as a measure to evaluate the consistency of ratings versus text. Our benchmarking reveals a significant performance gap: score-based models achieve a prediction accuracy of 91\%, whereas text-based models  stagnate at approximately 81\%. {This discrepancy and the existence of unpredictable samples raise two critical questions: a) What are the characteristics of the remaining 9\% {hard samples} that defy score-based prediction? b) Why does text-based prediction perform so much worse than score-based prediction?}

To answer the first question, we analyzed the statistical characteristics of the hard samples. Results show that the score distributions of these samples exhibit {high kurtosis} and {negative skewness}. This means that specific low scores play a decisive role in the rejection decision, even if the average score is near the acceptance threshold. Unlike simple variance, this negative asymmetry suggests that a strong objection often outweighs the consensus of other reviewers.

To answer the second question, we investigated the underlying cause of text ambiguity. We attribute the poor performance of text prediction to the Politeness Principle \cite{brown1987politeness}. Our aspect-based sentiment analysis shows that reviews of rejected papers still contain a higher proportion of positive sentiment words than negative ones. This polite expression masks the true rejection intention, making it difficult for authors to judge the outcome based on reading the text alone.

In summary, this work makes three key contributions. First, we release a processed dataset of multi-turn review dialogues. Second, we quantify the signal decoupling between scores and text. Third, we identify high kurtosis and negative skewness as the defining characteristics of hard samples, and verify the Politeness Principle as the primary cause of text ambiguity.

\section{Related work}
Peer review refers to the  evaluation of research manuscripts by independent experts in the same field. While extensive research has been conducted to model this process, previous works have predominantly adopted a {model-centric} perspective, aiming to maximize prediction accuracy through advanced architectures. In contrast, our work shifts to a {data-centric} view, investigating the intrinsic boundaries of predictability and the characteristics of samples that defy algorithmic assessment.

\subsection{Paper Acceptance Prediction}
The digitalization of peer review has spurred a wave of research focused on predicting paper acceptance. The establishment of benchmarks, such as the PeerRead dataset \cite{kang-etal-2018-dataset}, catalyzed this direction by treating acceptance prediction as a standard classification task. Subsequent research has largely been driven by the goal of improving performance metrics through feature engineering and complex model designs. 

Deep learning approaches have evolved from utilizing basic sentiment analysis \cite{10.1145/3209978.3210056} to fusing content and sentiment features via frameworks like DeepSentiPeer \cite{ghosal-etal-2019-deepsentipeer}. To capture longer dependencies and hierarchical structures, researchers have deployed hybrid models combining CNNs and Bi-LSTMs \cite{10.1007/978-3-030-86230-5_60}, as well as self-attention networks \cite{deng-etal-2020-hierarchical}. Beyond textual semantics, multimodal features such as visual layout have also been exploited to squeeze out marginal performance gains.

Recently, the advent of Large Language Models (LLMs) has further transformed this field. Applications range from argument generation \cite{hua-etal-2019-argument-generation} to collaborative review generation \cite{10.1145/3340555.3353766, wang-etal-2020-reviewrobot}. Specialized models like OpenReviewer \cite{idahl-ahmadi-2025-openreviewer} and multi-agent systems \cite{lu2025agent} have been developed to mimic human review processes. However, current research largely focuses on LLMs as {generators} rather than {discriminators} \cite{zhou-etal-2024-llm, ZHUANG2025103332}. Trust issues also persist, with findings of narcissistic bias where LLMs favor AI-generated text \cite{li2025llm} and susceptibility to indirect injection attacks \cite{Zhu2025WhenYR}.

Crucially, these works typically attribute prediction failures to model limitations or random noise. They rarely investigate whether a subset of data possesses {structural hardness} that renders it inherently unpredictable by standard computational means. Our study fills this gap by identifying challenging samples that remain misclassified regardless of model complexity.
\begin{table}[!ht] 
    \centering 
    \caption{Evolution of the semantic definitions for review ratings in ICLR (2021–2025).}
    \label{tab:rating-meaning} 
    \renewcommand{\arraystretch}{1.2} 
    \resizebox{\columnwidth}{!}{
    \begin{tabular}{c c l} 
        \toprule 
        \textbf{Year} & \textbf{Score} & \textbf{Meaning} \\
        \midrule
        
        \multirow{6}{*}{2022$\sim$2025} 
        & 1 & Strong reject \\
        & 3 & Reject \\
        & 5 & Marginally reject \\
        & 6 & Marginally accept \\
        & 8 & Accept \\
        & 10 & Strong accept \\
        \midrule 
        
        \multirow{10}{*}{2021}
        & 1 & Trivial or wrong \\
        & 2 & Strong rejection \\
        & 3 & Clear rejection \\
        & 4 & Ok but not good enough -- rejection \\
        & 5 & Marginally below acceptance threshold \\
        & 6 & Marginally above acceptance threshold \\
        & 7 & Good paper, accept \\
        & 8 & Top 50\% of accepted papers, clear accept \\
        & 9 & Top 15\% of accepted papers, strong accept \\
        & 10 & Top 5\% of accepted papers, seminal paper \\
        \bottomrule
    \end{tabular}
    }
\end{table}
%%%%%%%%%%%%%%%%%%%%%%%%%%%%%%%%%%%%%%
%%%%%%%%%%%%%%%%%%%%%%%%%%%%%%%%%%%%%%
\begin{figure*}[t]
    \centering
    \includegraphics[width=0.95\textwidth]{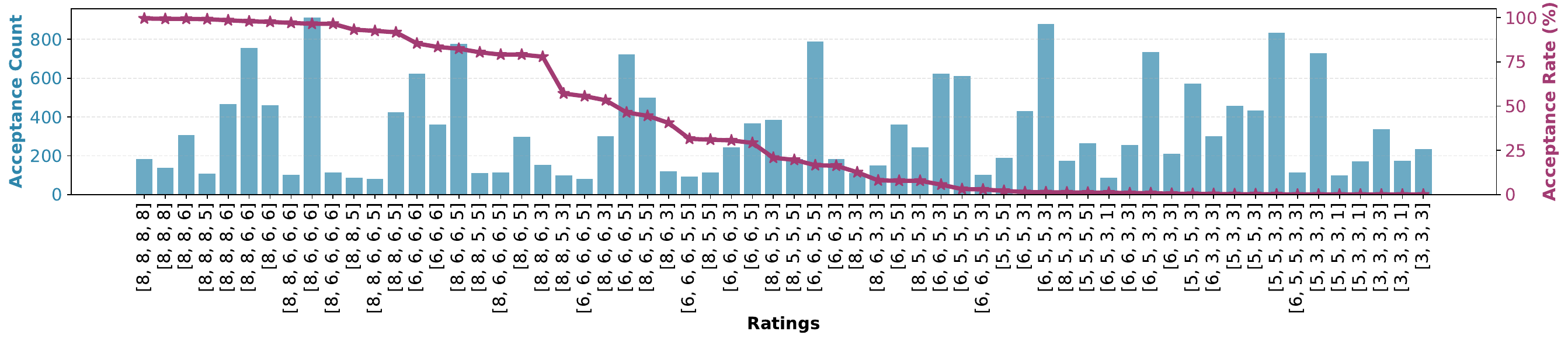} % 跨两栏时建议用 0.8-0.9 宽度
    \caption{The count and rate of accepted papers combined from 2021 to 2025.}
    \label{fig:rating_acceptance_analysis}
\end{figure*}
%%%%%%%%%%%%%%%%%%%%%%%%%%%%%%%%%%%%%%
%%%%%%%%%%%%%%%%%%%%%%%%%%%%%%%%%%%%%%
\begin{table*}[!t] 
\centering
\caption{Comparative statistics between official ICLR records and our processed dataset (2021--2025).}
\label{tab:statistic_info}
\small
\renewcommand{\arraystretch}{1.05}
\setlength{\tabcolsep}{12pt} 

\begin{tabular}{llccccc}
\toprule
\textbf{Category} & \textbf{Metric} & \textbf{2021} & \textbf{2022} & \textbf{2023} & \textbf{2024} & \textbf{2025} \\
\midrule

\multirow{3}{*}{\textbf{Official Data}} 
& Submission & 3014 & 3391 & 4938 & 7262 & 11,603 \\
& Accept & 1027 & 1095 & 1574 & 2260 & 3704 \\
& Accept rate & 34.07\% & 32.26\% & 31\% & 31\% & 32\% \\
\midrule

\multirow{6}{*}{\textbf{Crawled Data}} 
& Submission & 3014 & 3422 & 4955 & 7404 & 11,672 \\
& Accept & 859 & 1094 & 1573 & 2260 & 3704 \\
& Reject & 1729 & 1523 & 2220 & 3439 & 4942 \\
& Withdraw & 403 & 779 & 1144 & 1652 & 2956 \\
& Desk reject & 17 & 26 & 18 & 53 & 70 \\
& Accept rate & 28.5\% & 31.97\% & 31.75\% & 30.52\% & 31.73\% \\
\midrule

\multirow{4}{*}{\textbf{Our Data}} 
& Submission & 2972 & 3354 & 4897 & 7210 & 11,512 \\
& Accept & 859 & 1084 & 1561 & 2248 & 3699 \\
& Labeled reject & 2113 & 2270 & 3336 & 4962 & 7813 \\
& Accept rate & 28.9\% & 32.32\% & 31.88\% & 31.18\% & 32.13\% \\
\bottomrule
\end{tabular}
\end{table*}
%%%%%%%%%%%%%%%%%%%%%%%%%%%%%%%%%%%%%%

\subsection{Linguistic Patterns in Peer Review}
Beyond static prediction, understanding the dynamics of the review lifecycle is crucial. Recent surveys highlight the complexity of author-reviewer interactions \cite{gao-etal-2019-rebuttal,HUANG2023101427}. Studies have tracked score trajectories, finding that while rebuttals act as a game changer for borderline papers \cite{fernandes2024enhancing, 10.1145/3529372.3530935}, social factors like peer pressure (herding) \cite{gao-etal-2019-rebuttal} and reviewer activity patterns \cite{kargaran2025insights} significantly influence the outcome.

A more subtle challenge impeding the alignment of text and decisions is the Politeness Principle. Linguistic studies have quantified the prevalence of polite markers masking harsh sentiments \cite{10.1007/s10579-023-09662-3}. While previous works acknowledge this as a linguistic feature, they have not systematically quantified its impact on the {signal decoupling} between scores and text. Our work demonstrates that this polite ambiguity disproportionately affects the reliability of text-based models compared to score-based baselines.
% arsh sentiments \cite{verma-etal-2022-lack, 10.1007/s10579-023-09662-3}. 

\section{Dataset Construction: ICLR 2021-2025}

\subsection{Data Collection and Dialogue Reconstruction}
We constructed a comprehensive dataset spanning the ICLR conferences from 2021 to 2025. All data were systematically collected from the official OpenReview platform via the Python API to ensure source authenticity. The dataset is organized at the paper level, with metadata encompassing the forum ID, title, abstract, author list, and keywords. For review content, we collected detailed comments, confidence scores, and overall ratings. It is important to note that the semantic meaning of ratings evolved over time. Table \ref{tab:rating-meaning} details the specific definitions for each score across different years. We utilized the raw numeric values for analysis.

To ensure data quality, we implemented a rigorous cleaning and reconstruction pipeline. First, we excluded submissions labeled as {Desk Rejected} or {Withdrawn} prior to receiving reviews, as they lack core interaction data. Second, we removed Public Comments and cross-replies to strictly confine the study to formal interactions between designated reviewers and authors. We also removed procedural metadata such as code of conduct acknowledgments.

A critical step in our pipeline is the structuring of raw comments into independent multi-round review sessions. Unlike previous datasets that may aggregate comments loosely, we leveraged the \texttt{replyto} field to trace the exact relational structure of the interactions. We grouped all content originating from a specific reviewer's initial review into a dedicated session. Within each session, we sorted the exchanges strictly in chronological order. Consequently, each paper is represented by multiple distinct interaction sequences, capturing the full dialogue flow. Finally, we standardized the decision labels by consolidating all acceptance-related categories into a single {Accept} class and classifying post-review withdrawals as {Reject}, yielding a binary classification task.

\subsection{Statistics and Consistency Check}
To validate the integrity of our dataset, we compared our processed statistics with the official ICLR records. Table \ref{tab:statistic_info} presents a detailed comparison of submission counts and acceptance rates from 2021 to 2025. The data exhibits a high degree of consistency. The discrepancy between the official data and our preprocessed dataset is minimal, typically ranging between 0.7\% and 1.4\% for submission counts. This slight difference is an expected result of our filtering procedure, which removes papers that did not complete the full peer-review process.
Figure \ref{fig:rating_acceptance_analysis} illustrates the distribution of ratings over the five-year period. This breakdown reveals the shifting  trends in scoring patterns and provides a statistical foundation for the subsequent predictability analysis. Additionally, Figure \ref{fig:rating_acceptance_analysis} visualizes the relationship between  score combinations and acceptance rates.

% \begin{table*}[t] 
%     \centering
%     \caption{Distribution of review ratings across five years (2021--2025), illustrating the shift in scoring trends.}
%     \label{tab:rating_distribution}
%     \small 
%     \renewcommand{\arraystretch}{1.2} 
%     % \setlength{\tabcolsep}{7pt}
%     \begin{tabular}{l r r r r r r r r r r r}
%         \toprule
%         \textbf{Year} & \textbf{1} & \textbf{2} & \textbf{3} & \textbf{4} & \textbf{5} & \textbf{6} & \textbf{7} & \textbf{8} & \textbf{9} & \textbf{10} & \textbf{Total} \\
%         \midrule
%         \textbf{2021} & 10 & 117 & 652 & 1746 & 2248 & 2669 & 1947 & 489 & 111 & 10 & 9,999 \\
%         \textbf{2022} & 156 & --- & 1908 & --- & 2698 & 3300 & --- & 2007 & --- & 50 & 10,119 \\
%         \textbf{2023} & 263 & --- & 2795 & --- & 3618 & 4772 & --- & 2722 & --- & 97 & 14,267 \\
%         \textbf{2024} & 612 & --- & 6939 & --- & 8050 & 8280 & --- & 3855 & --- & 102 & 27,838 \\
%         \textbf{2025} & 1022 & --- & 11459 & --- & 13019 & 14610 & --- & 6151 & --- & 164 & 46,425 \\
%         \bottomrule
%     \end{tabular}
% \end{table*}

% \clearpage

%%%%%%%%%%%%%%%%%%%%%%%%%%%%%%%%%%%%%%%%%%%%
\begin{table}[!t]
    \centering
    \caption{Acceptance prediction with rating.}
    \label{tab:main_results_rating}
    \renewcommand{\arraystretch}{0.95}
    \resizebox{0.85\columnwidth}{!}{%
    \begin{tabular}{l l l}
        \toprule
        \textbf{Input} & \textbf{Model} & {\textbf{Accuracy}} \\
        \midrule
        
        \multirow{11}{*}{{Rating-Only}} 
        & Qwen-turbo & 0.7600\\
        & Claude-haiku-4-5 & 0.8100\\
        & Gemini-2.5 pro & 0.8400\\
        & GPT-5 & 0.8600 \\
        & SVM  & 0.9045 \\
        &  Threshold & 0.9065 \\
        & LR & 0.9067 \\
        & Random Forest & 0.9068 \\
        & XGBoost & 0.9099 \\
        & MLP & 0.9100 \\
        \cmidrule(l){2-3}
        & \textit{Average}& 0.8714 \\
        \bottomrule
    \end{tabular}%
    }
\end{table}
%%%%%%%%%%%%%%%%%%%%%%%%%%%%%%%%%%%%%%%%%%%%
%%%%%%%%%%%%%%%%%%%%%%%%%%%%%%%%%%%%%%%%%%%%
\begin{table}[!t]
    \centering
    \caption{Acceptance prediction with review.}
    \label{tab:main_results_text}
    \renewcommand{\arraystretch}{1.0}
    \resizebox{0.9\columnwidth}{!}{%
    \begin{tabular}{l l l}
        \toprule
        \textbf{Input} & \textbf{Model} & {\textbf{Accuracy}} \\
        \midrule
        \multirow{8}{*}{{Initial Review}}
        & Word2Vec & 0.6595 \\
        & Qwen-turbo & 0.6700\\
        & TF-IDF  & 0.7061 \\
        & TextCNN & 0.7128 \\
        & GPT-5 & 0.7400 \\
        & Gemini-2.5 pro & 0.7600\\
        & Claude-haiku-4-5 & 0.7800\\
        \cmidrule(l){2-3}
        & \textit{Average} & 0.7183 \\
        \midrule

        \multirow{9}{*}{{Weakness}}
        & Word2Vec & 0.6679 \\
        & TF-IDF  & 0.6974 \\
        & SciBert & 0.7120 \\
        & GPT-5 & 0.7200 \\
        & Claude-haiku-4-5 & 0.7200\\
        & Qwen-turbo & 0.7300\\
        & TextCNN & 0.7307 \\
        & Gemini-2.5 pro & 0.8100\\
        \cmidrule(l){2-3}
        & \textit{Average} & 0.7235 \\
        \midrule

        \multirow{8}{*}{{Strength+Weakness}}
        & Word2Vec & 0.6682 \\
        & TextCNN & 0.6911 \\
        & TF-IDF  & 0.7121 \\
        & Claude-haiku-4-5 & 0.7400\\
        & GPT-5 & 0.7800 \\
        & Qwen-turbo & 0.7800\\
        & Gemini-2.5 pro & 0.7800\\
        \cmidrule(l){2-3}
        & \textit{Average} & 0.7359 \\
        \bottomrule
    \end{tabular}%
    }
\end{table}
%%%%%%%%%%%%%%%%%%%%%%%%%%%%%%%%%%%%%%%%%%%%

%%%%%%%%%%%%%%%%%%%%%%%%%%%%%%%%%%%%%%%%%%%%
\begin{table}[!t]
    \centering
    \caption{Acceptance prediction with rating and review.}
    \label{tab:main_results_both}
    \renewcommand{\arraystretch}{1.0}
    \resizebox{1\columnwidth}{!}{%
    \begin{tabular}{l l l}
        \toprule
        \textbf{Input} & \textbf{Model} & {\textbf{Accuracy}} \\
        \midrule
        
        \multirow{6}{*}{{Rating+Review}} 
        & Qwen-turbo & 0.7500\\
        & Dual-Branch Attention & 0.8107 \\
        & GPT-5 & 0.8300 \\
        & Claude-haiku-4-5 & 0.8400\\
        & Gemini-2.5 pro & 0.8600\\
        \cmidrule(l){2-3}
        & \textit{Average} & 0.8181 \\
        \midrule
        
        \multirow{6}{*}{{Rating+Weakness}} 
        & Gemini-2.5 pro & 0.7900\\
        & Qwen-turbo & 0.8000\\
        & Dual-Branch Attention & 0.8122 \\
        & GPT-5 & 0.8500 \\
        & Claude-haiku-4-5 & 0.8500\\
        \cmidrule(l){2-3}
        & \textit{Average}& 0.8004 \\
        \midrule

        \multirow{6}{*}{{Rating+Weak+Stre}} 
        & Qwen-turbo & 0.5900\\
        & Claude-haiku-4-5 & 0.8200\\
        & Gemini-2.5 pro & 0.8500\\
        & GPT-5 & 0.8600 \\
        & Dual-Branch Attention & 0.8682 \\
        \cmidrule(l){2-3}
        & \textit{Average} & 0.7976 \\
        \midrule

        \multirow{2}{*}{{Rating+Rebuttal}} 
        & Dual-Branch Attention & 0.8258 \\
        \cmidrule(l){2-3}
        & \textit{Average} & 0.8258 \\

        \bottomrule
    \end{tabular}%
    }
\end{table}
%%%%%%%%%%%%%%%%%%%%%%%%%%%%%%%%%%%%%%%%%%%%%

\section{Benchmarking and Signal Decoupling}

%%%%%%%%%%%%%%%%%%%%%%%%%%%%%%%%%%%%%%%%%%%%%%%%
\begin{figure*}[!t]
    \centering
    \includegraphics[width=0.83\textwidth]{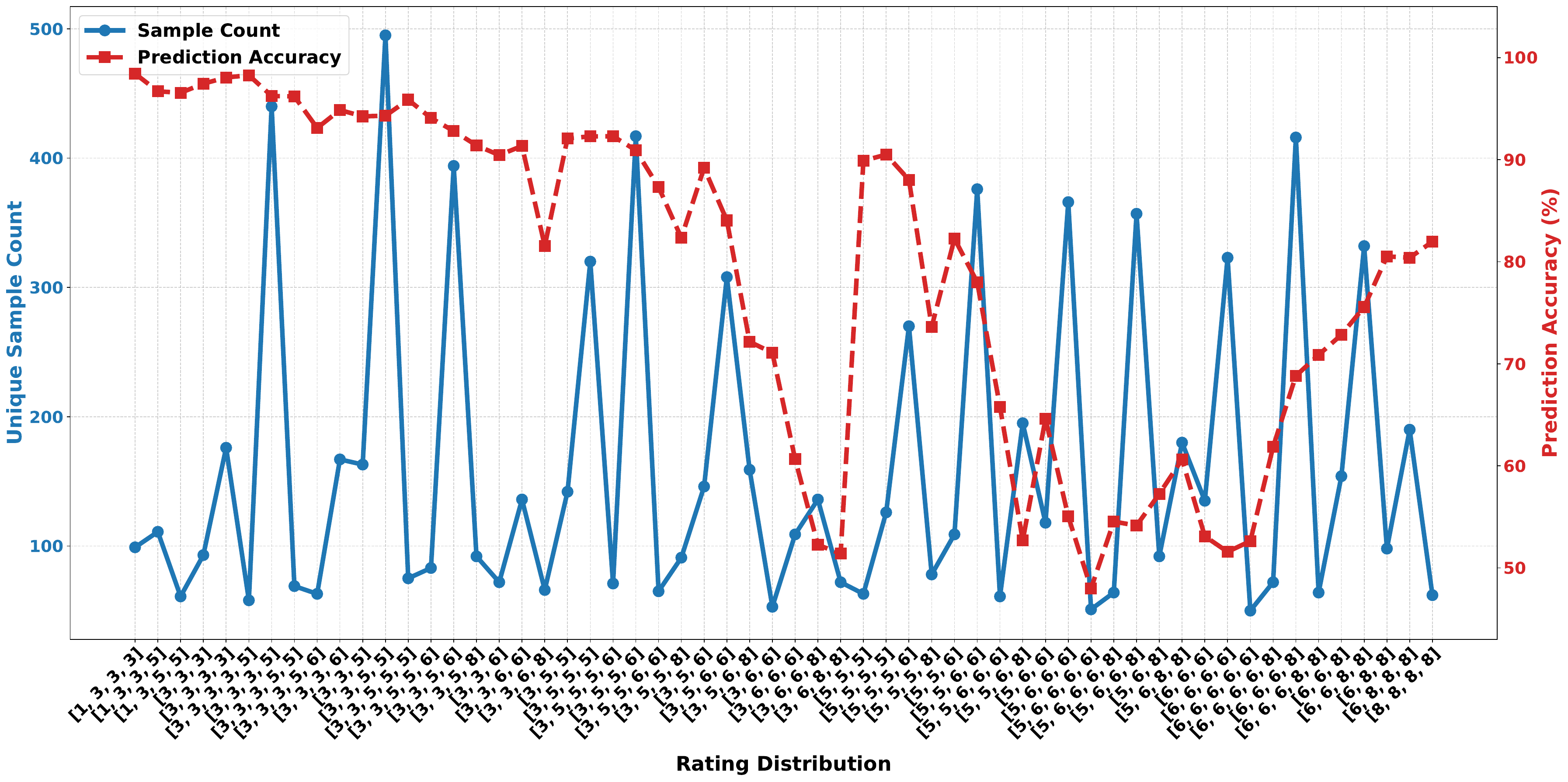}
    \caption{Prediction accuracy across different rating, arranged in ascending order of average score. The trend exhibits a distinct {U-shaped structure}: accuracy is initially high for consensus rejections, significantly deteriorates to near-random guessing ($\sim$50\%) for borderline cases with conflicting scores, and subsequently rebounds for high-scoring acceptances.}
    \label{fig:AccuracyDistritbuion}
\end{figure*}
%%%%%%%%%%%%%%%%%%%%%%%%%%%%%%%%%%%%%%%%%%%%%%%%
\subsection{Experimental Setup}
We frame acceptance prediction as a binary classification task. To prevent data leakage, we employ a temporal split: submissions from 2022–2024 constitute the training set, while the 2025 cohort serves as the hold-out test set ($N=10,512$). We benchmark three modeling paradigms: (1) Traditional ML, including Random Forest, XGBoost, and SVM,  (2) Deep Learning, utilizing Word2Vec, TextCNN, and SciBERT \cite{beltagy-etal-2019-scibert} (restricted to the ``Weakness'' section due to context length),  and (3) LLMs, specifically Qwen-Turbo, Gemini-2.5-pro, GPT-5, and Claude-Haiku-4.5, evaluated in a zero-shot meta-reviewer setting. We also conducted an ablation study across Rating-Only, Text-Only, and Hybrid settings. 
% Full hyperparameters are detailed in the Appendix.
%  Word2Vec \cite{Mikolov2013EfficientEO}, TextCNN \cite{kim-2014-convolutional}, and SciBERT \cite{beltagy-etal-2019-scibert} 

We benchmarked three distinct modeling paradigms with the following configurations. First, we evaluated state-of-the-art proprietary models including {Qwen-turbo}, {Claude-haiku-4.5}, {Gemini-2.5 Pro}, and {GPT-5} in a zero-shot setting. These models were accessed via their respective official APIs using default decoding parameters to simulate a standard meta-reviewer scenario. 

To ensure reproducibility, the specific prompt template designed for the rating-based prediction task is: {Drawing upon your extensive knowledge of ICLR conference standards and historical acceptance trends, predict the final decision for this paper based solely on the provided ratings: [YOUR RATINGS]. The scoring scale is: 10: Strong Accept; 8: Accept; 6: Marginally Accept; 5: Marginally Reject; 3: Reject; 1: Strong Reject. You must output the result strictly in the following format: Decision: [Accept or Reject]. Do not provide any reasons or explanations.}

Second, baseline models were implemented using Scikit-learn. The {SVM} utilized a linear kernel with $C=10$. Ensemble methods were calibrated for robustness: {Random Forest} was set to 200 estimators with a minimum split sample of 10, while {XGBoost} employed a learning rate of 0.05 and a maximum depth of 3 to prevent overfitting. For semantic baselines, we used Logistic Regression ($C=5$), and trained {Word2Vec} embeddings ($d=200$) fed into an MLP classifier. Additionally, a simple heuristic baseline was established using an Average Rating Threshold of 5.8.
Then, for neural architectures, {TextCNN} was configured with an embedding dimension of 300 and 150 filters. The pre-trained {SciBERT} model was fine-tuned with a learning rate of $1.76 \times 10^{-5}$ and a batch size of 16. The {Dual-Branch Attention} network used a learning rate of $9.35 \times 10^{-6}$ and a dropout rate of 0.104.

\subsection{The Decoupling Phenomenon}
Our experiments reveal a striking divergence between the predictive power of quantitative ratings and qualitative text. 
 As shown in Table \ref{tab:main_results_rating}, traditional ML models achieve the highest performance, with MLP reaching a ceiling of 91.00\%. Notably, a simple heuristic (Average Rating Threshold $> 5.8$) alone yields an accuracy of 90.65\%. This suggests that the decision boundary in the numerical domain is highly linear and explicit.

 A significant performance gap emerges when shifting to text-based prediction. As detailed in Table \ref{tab:main_results_text}, even state-of-the-art LLMs stagnate at approximately 81.00\% accuracy (using the ``Weakness'' section). Comparing Table \ref{tab:main_results_rating} and Table \ref{tab:main_results_both} highlights a persistent gap of approx. 10\% between the best rating-based model and the best text-based model. Even with multimodal fusion (Table \ref{tab:main_results_both}), accuracy peaks at 86.82\%, failing to surpass the rating-only baseline. This discrepancy confirms the {Signal Decoupling} phenomenon: numerical scores serve as the precise proxy for final decisions, whereas textual reviews contain substantial noise that obscures the ground truth.

 However, the dominance of scores is not absolute. While overall accuracy is high, we observe that prediction reliability is highly non-uniform. Specifically, the model certainty collapses for submissions with conflicting scores. As visualized in Figure \ref{fig:AccuracyDistritbuion}, the prediction accuracy for these {borderline cases} approaches random guessing ($\sim$50\%). This suggests the existence of {structural hardness}, a subset of samples that defy standard algorithmic rules. We define these as ``Hard Samples'' and pursue a fine-grained diagnosis of their characteristics in the next section.

%%%%%%%%%%%%%%%%%%%%%%%%%%%%%%%%%%%%%%%%%%%%%%%%
% \begin{figure*}[t] 
%     \centering
    
%     \begin{subfigure}[b]{0.32\textwidth}
%         \centering
%         \includegraphics[width=\textwidth]{Hard_Sample_rating_distribution.pdf}
%     \end{subfigure}
%     \hfill
%     \begin{subfigure}[b]{0.32\textwidth}
%         \centering
%         \includegraphics[width=\textwidth]{Simple_Accept_Sample_rating_distribution.pdf}
%     \end{subfigure}
%     \hfill
%     \begin{subfigure}[b]{0.32\textwidth}
%         \centering
%         \includegraphics[width=\textwidth]{Simple_Reject_Sample_rating_distribution.pdf}
%     \end{subfigure}

%     % \vspace{0.5cm} 

%     % 第二行
%     \begin{subfigure}[b]{0.32\textwidth}
%         \centering
%         \includegraphics[width=\textwidth]{Hard_Sample_confidence_distribution.pdf}
%     \end{subfigure}
%     \hfill
%     \begin{subfigure}[b]{0.32\textwidth}
%         \centering
%         \includegraphics[width=\textwidth]{Simple_Accept_Sample_confidence_distribution.pdf}
%     \end{subfigure}
%     \hfill
%     \begin{subfigure}[b]{0.32\textwidth}
%         \centering
%         \includegraphics[width=\textwidth]{Simple_Reject_Sample_confidence_distribution.pdf}
%     \end{subfigure}

%     \caption{Comparison of mean review ratings (top) and  prediction confidence (bottom) across hard samples, simples samples (including accept and reject).}
%     \label{fig:score_and_confidence}
% \end{figure*}
%%%%%%%%%%%%%%%%%%%%%%%%%%%%%%%%

%%%%%%%%%%%%%%%%%%%%%%%%%%%%%%%%%
\begin{figure*}[t]
    \centering
    % 第一行：Hard Sample
    \begin{subfigure}[b]{0.19\textwidth}
        \centering
        \includegraphics[width=\textwidth]{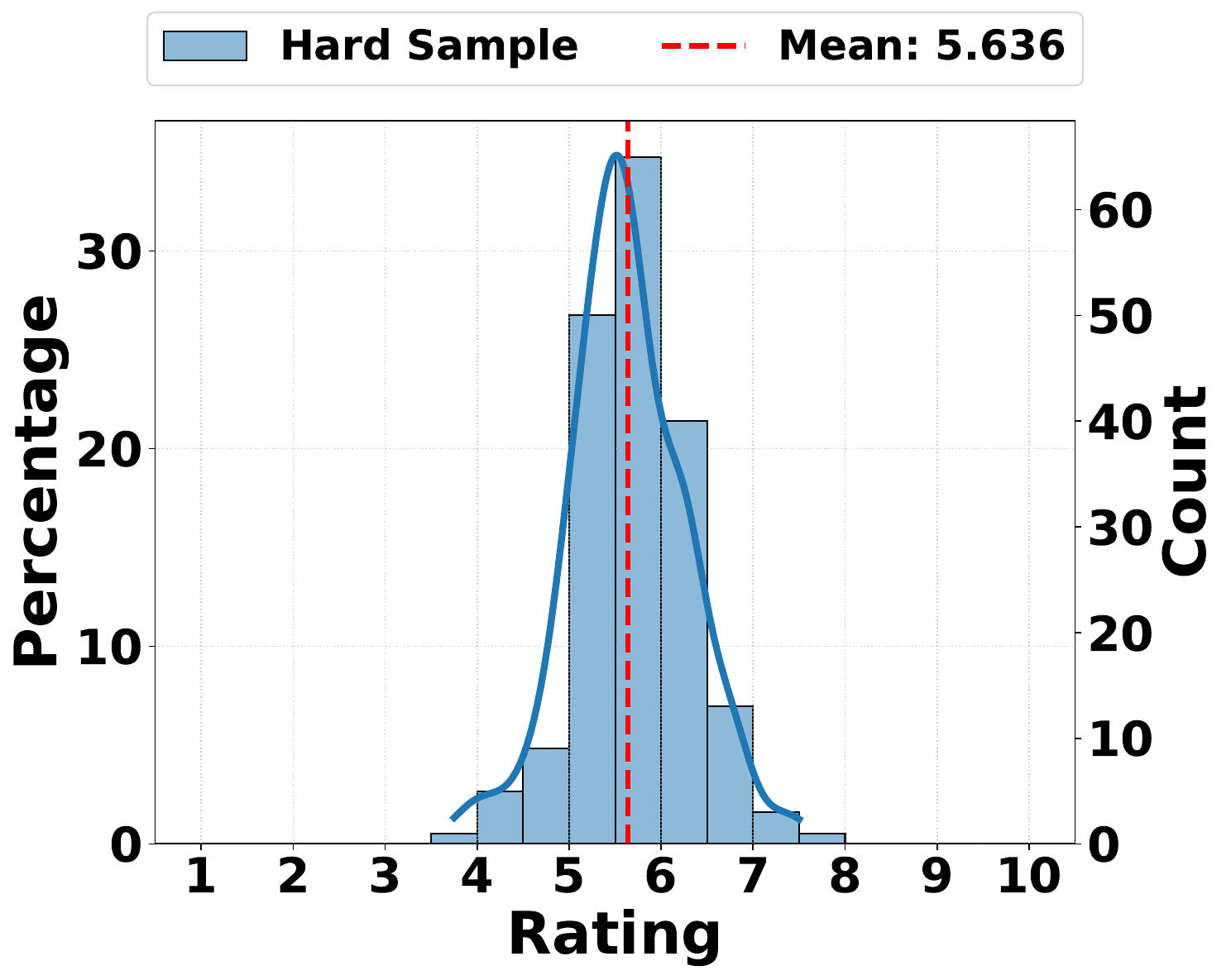}
    \end{subfigure}
    \hfill
    \begin{subfigure}[b]{0.19\textwidth}
        \centering
        \includegraphics[width=\textwidth]{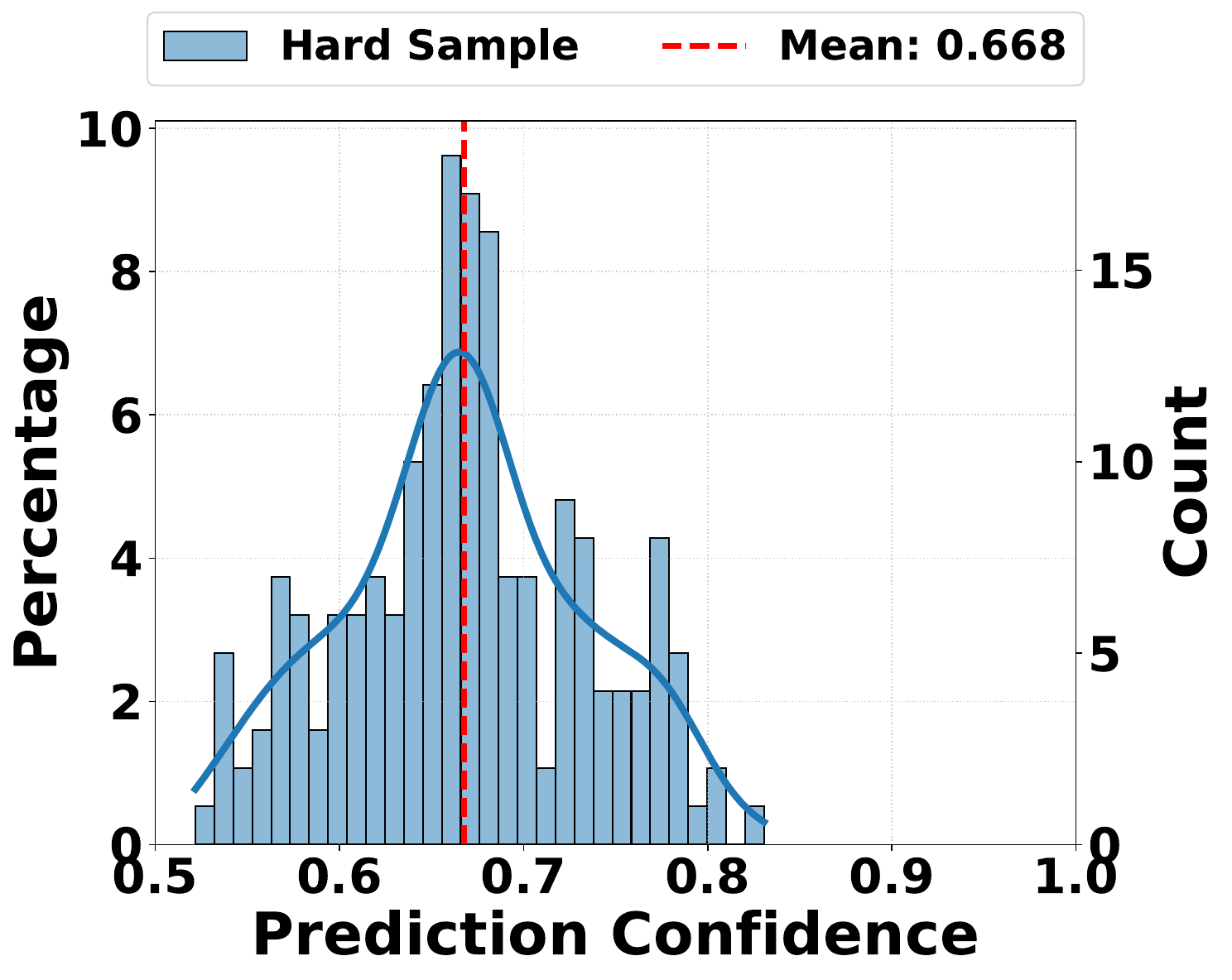}
    \end{subfigure}
    \hfill
    \begin{subfigure}[b]{0.19\textwidth}
        \centering
        \includegraphics[width=\textwidth]{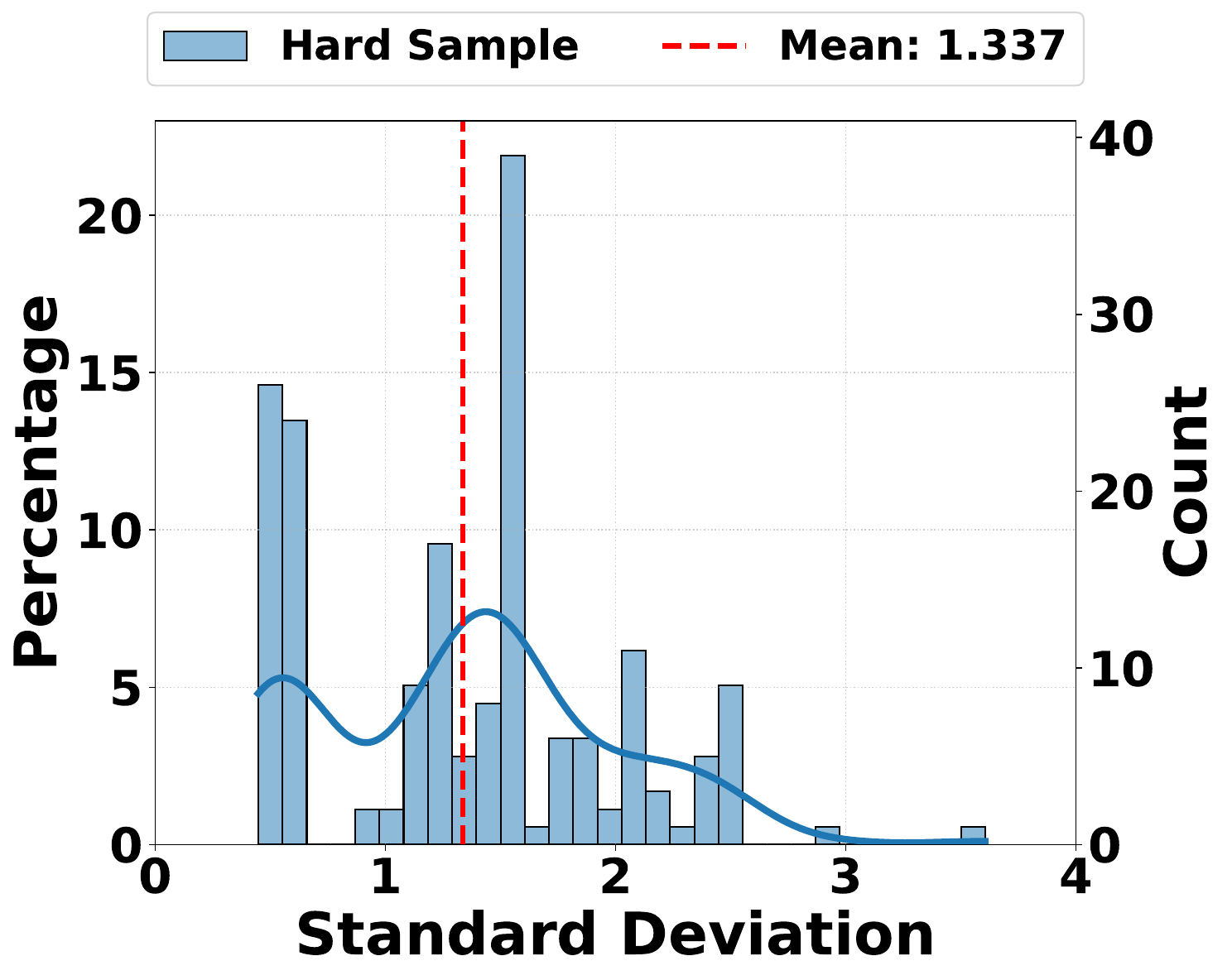}
    \end{subfigure}
    \hfill
    \begin{subfigure}[b]{0.19\textwidth}
        \centering
        \includegraphics[width=\textwidth]{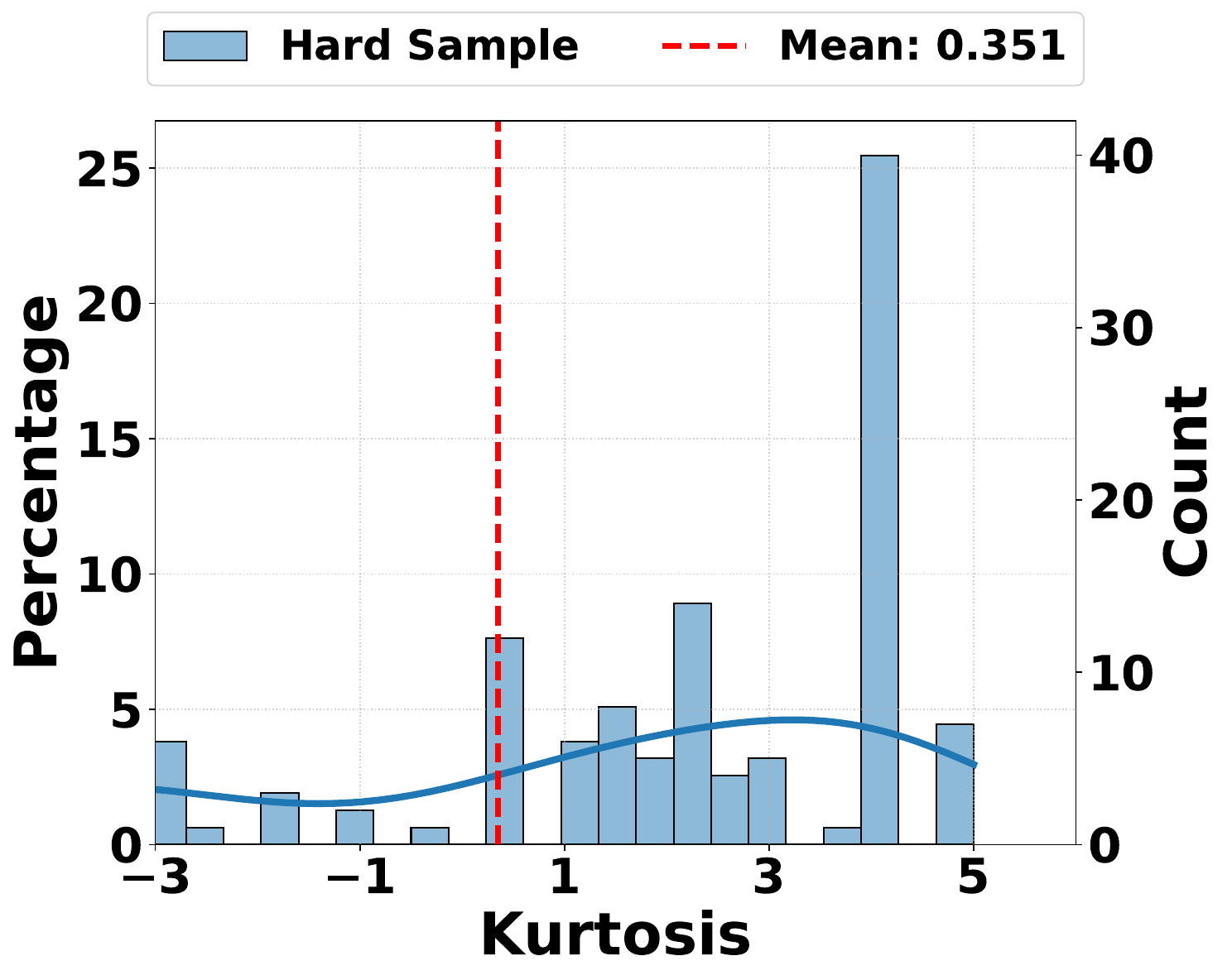}
    \end{subfigure}
    \hfill
    \begin{subfigure}[b]{0.19\textwidth}
        \centering
        \includegraphics[width=\textwidth]{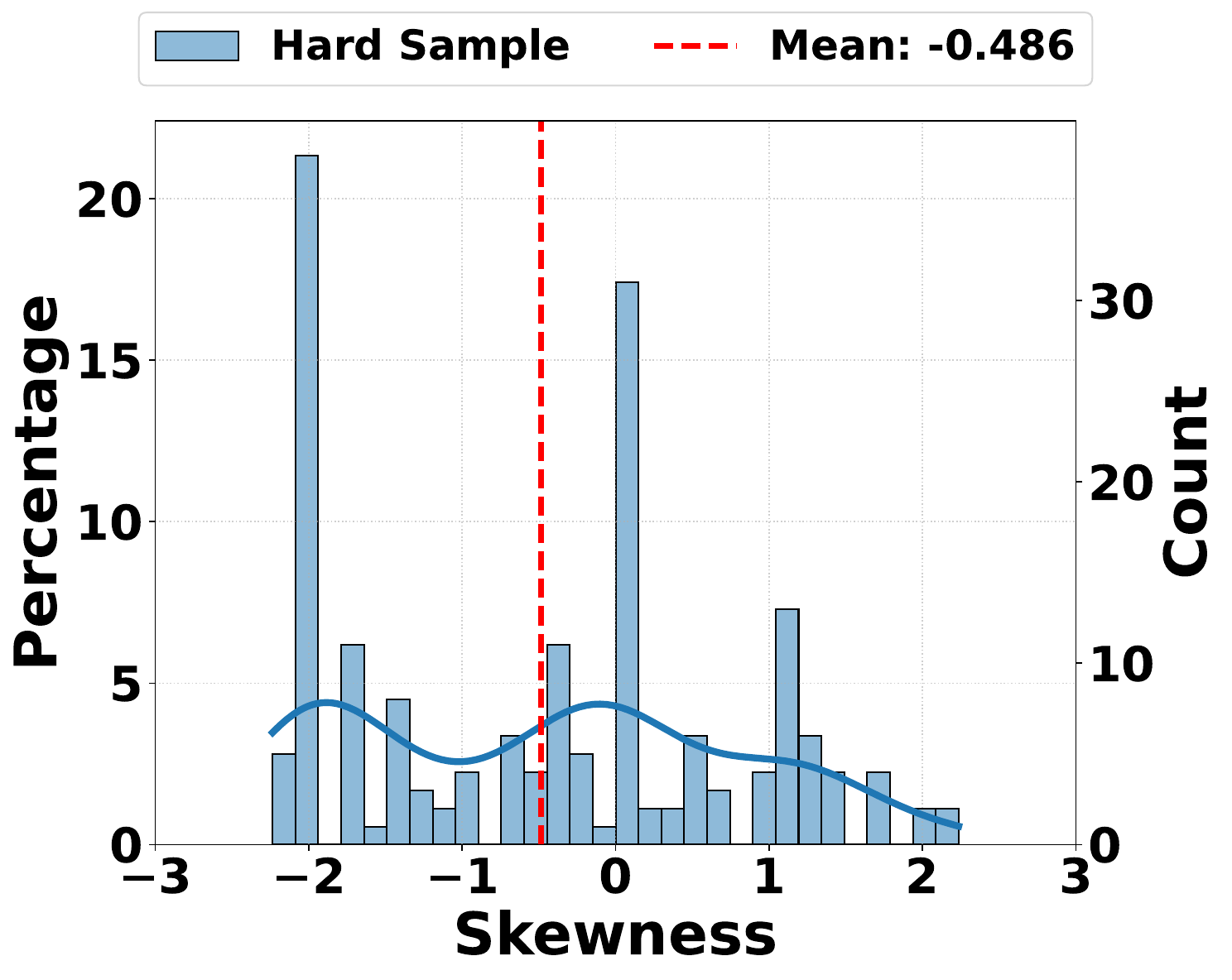}
    \end{subfigure}
    
    \vspace{0.2cm}
    % 第二行：Simple Accept Sample
    \begin{subfigure}[b]{0.19\textwidth}
        \centering
        \includegraphics[width=\textwidth]{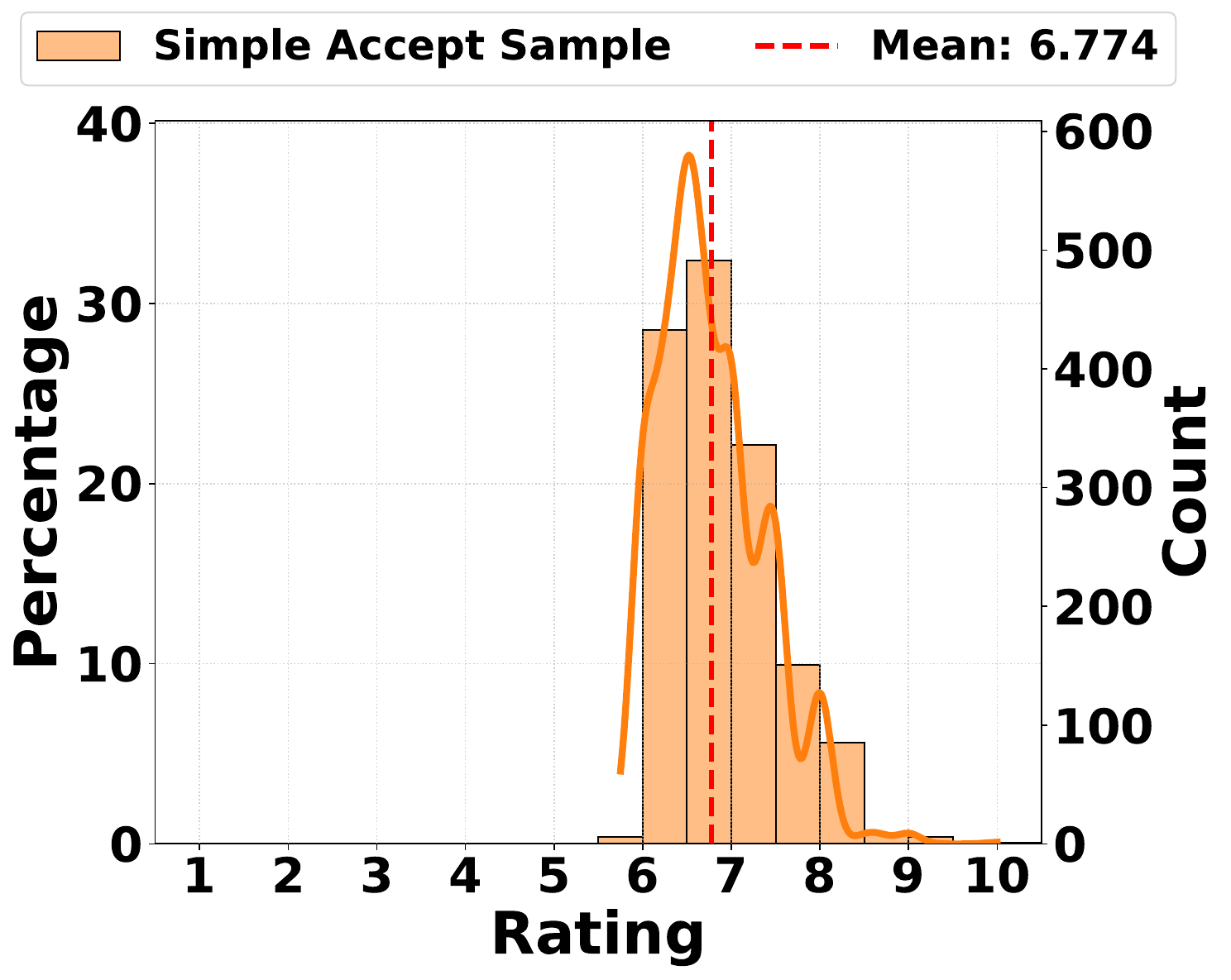}
    \end{subfigure}
    \hfill
    \begin{subfigure}[b]{0.19\textwidth}
        \centering
        \includegraphics[width=\textwidth]{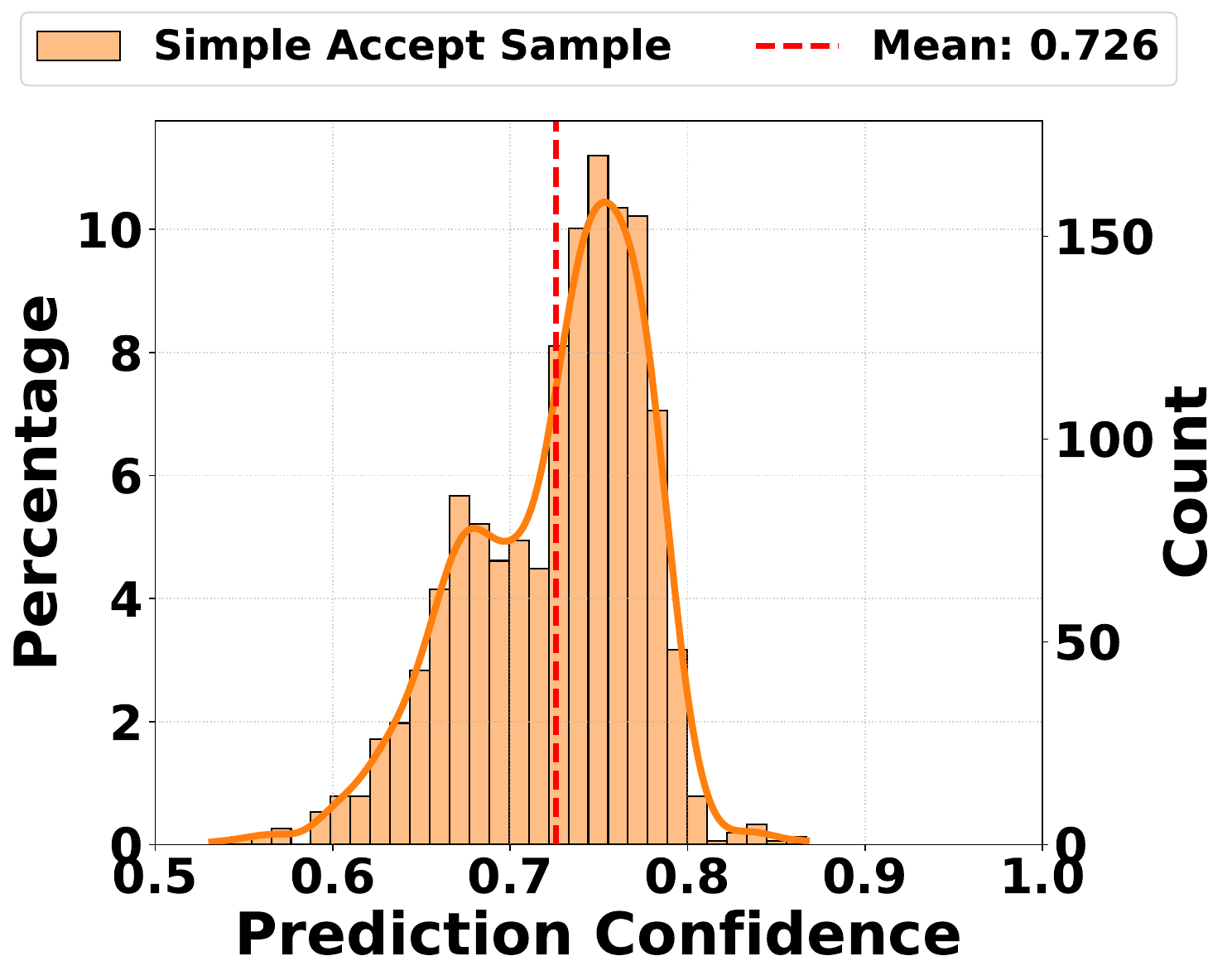}
    \end{subfigure}
    \hfill
    \begin{subfigure}[b]{0.19\textwidth}
        \centering
        \includegraphics[width=\textwidth]{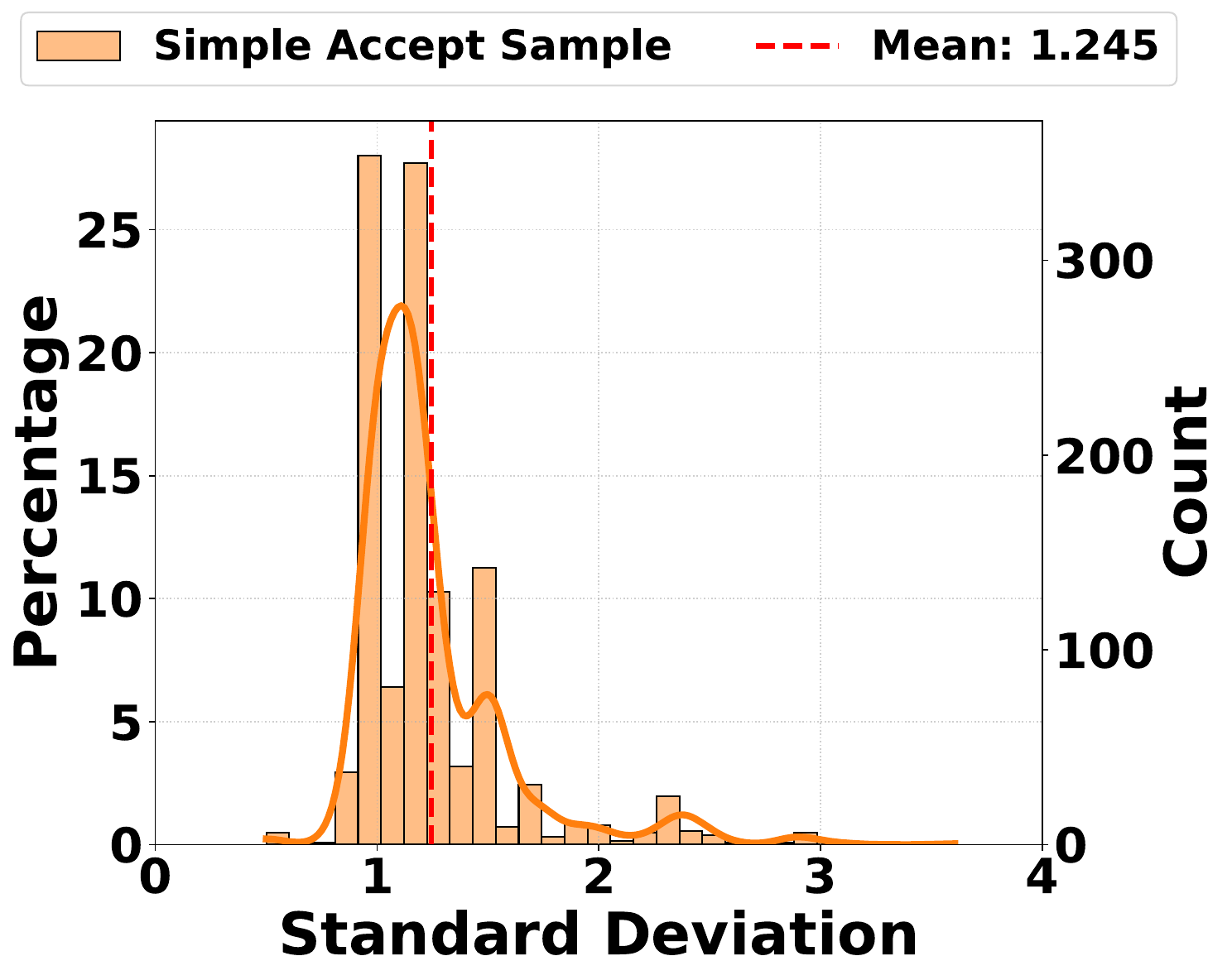}
    \end{subfigure}
    \hfill
    \begin{subfigure}[b]{0.19\textwidth}
        \centering
        \includegraphics[width=\textwidth]{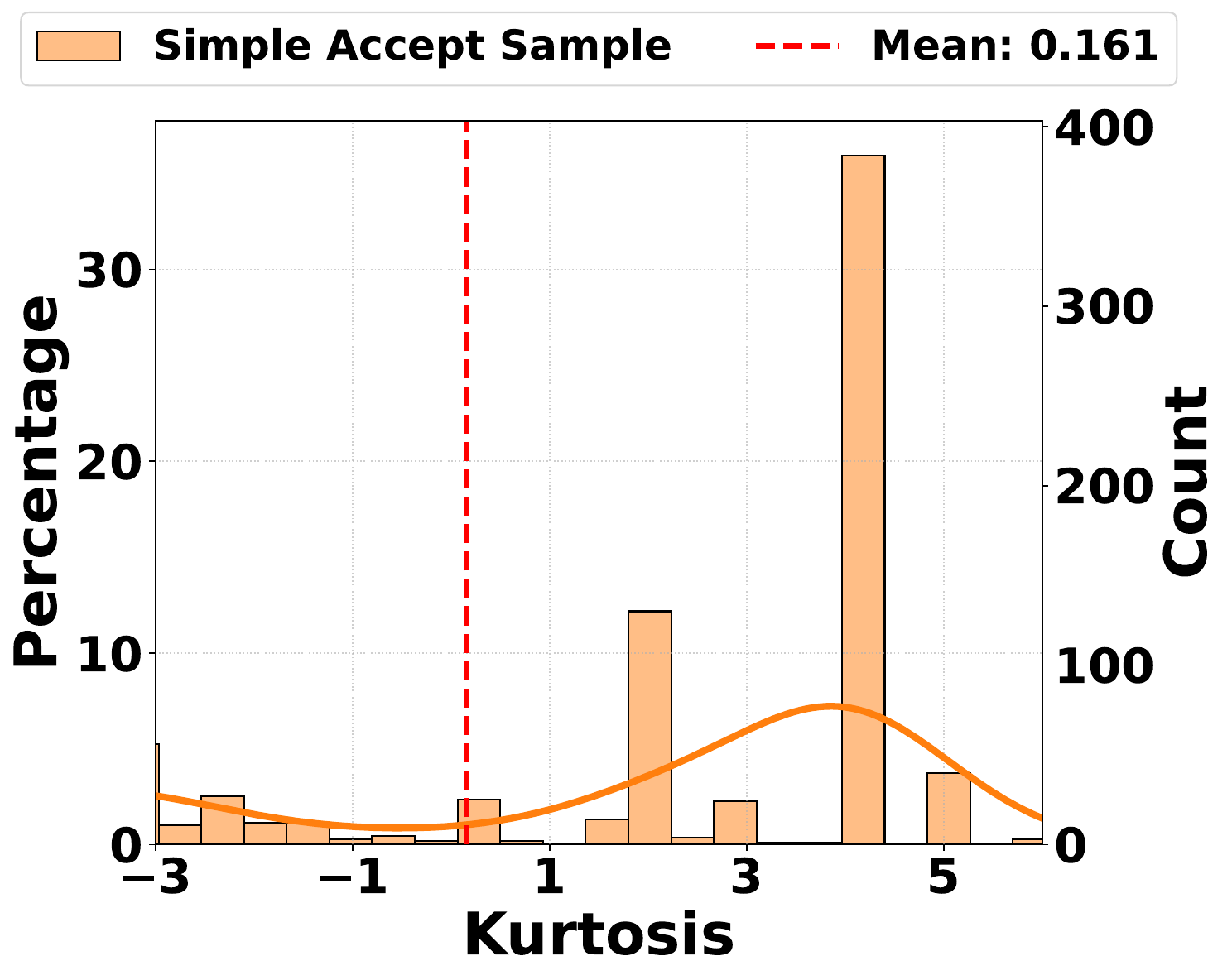}
    \end{subfigure}
    \hfill
    \begin{subfigure}[b]{0.19\textwidth}
        \centering
        \includegraphics[width=\textwidth]{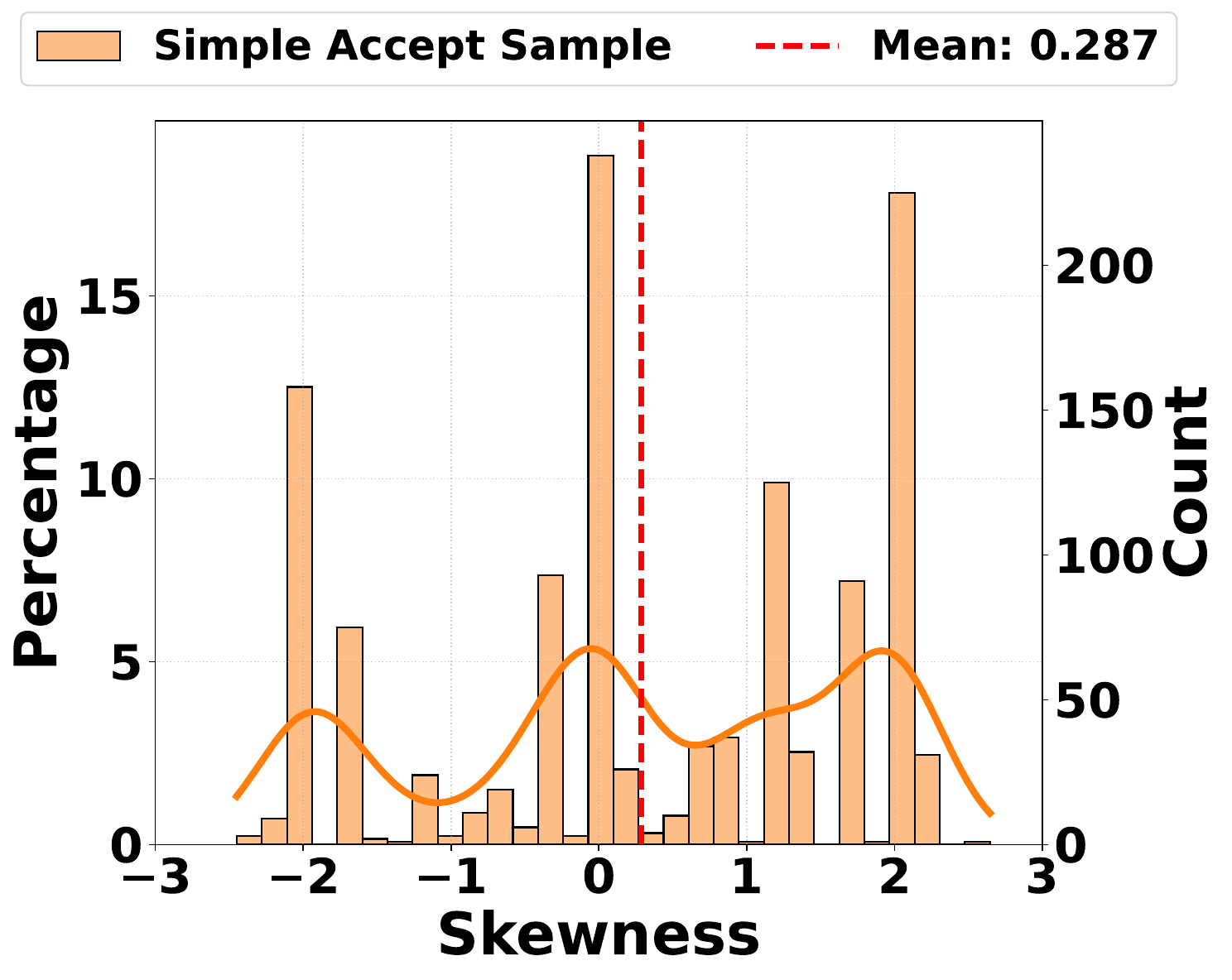}
    \end{subfigure}
    
    \vspace{0.2cm}
    % 第三行：Simple Reject Sample
    \begin{subfigure}[b]{0.19\textwidth}
        \centering
        \includegraphics[width=\textwidth]{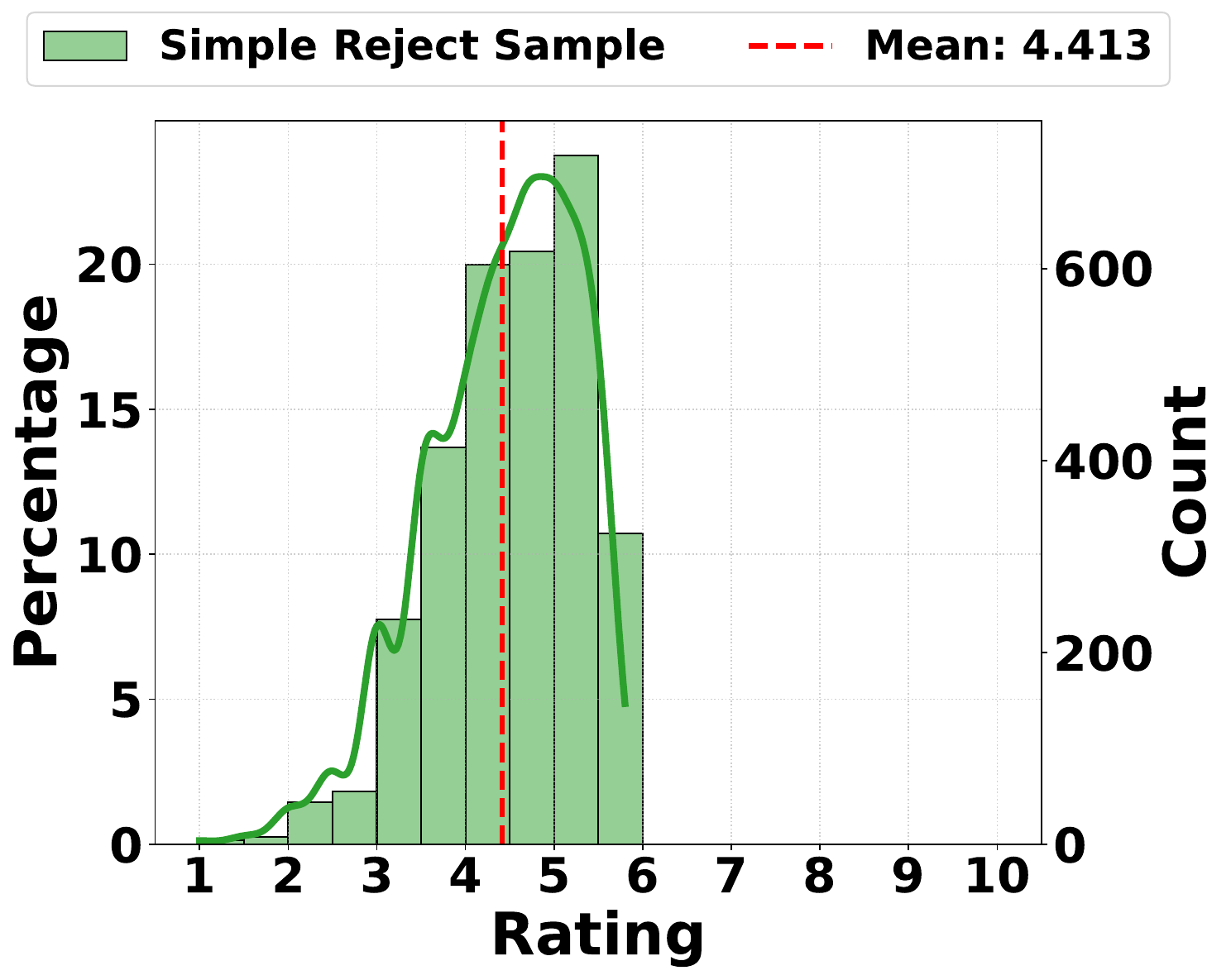}
    \end{subfigure}
    \hfill
    \begin{subfigure}[b]{0.19\textwidth}
        \centering
        \includegraphics[width=\textwidth]{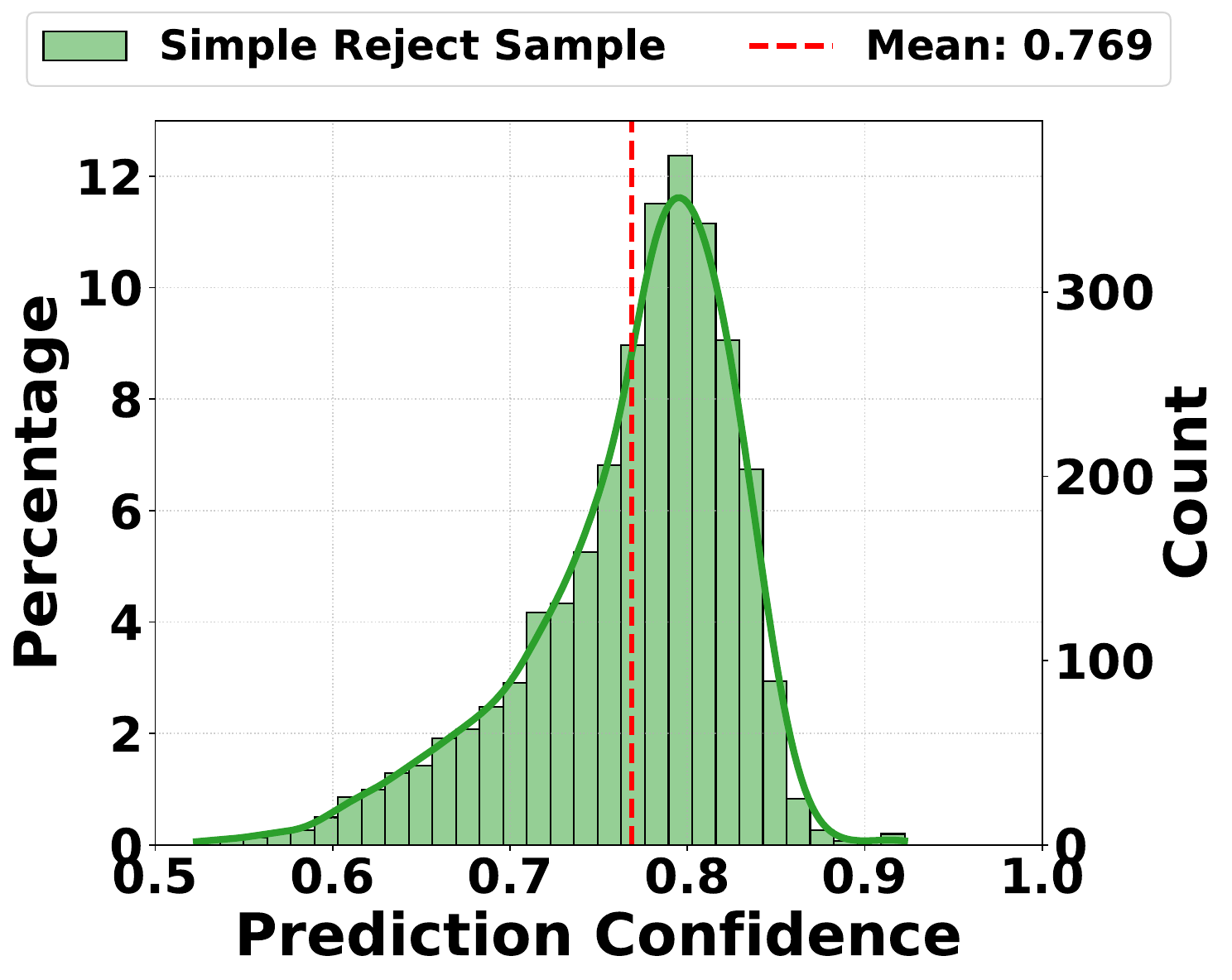}
    \end{subfigure}
    \hfill
    \begin{subfigure}[b]{0.19\textwidth}
        \centering
        \includegraphics[width=\textwidth]{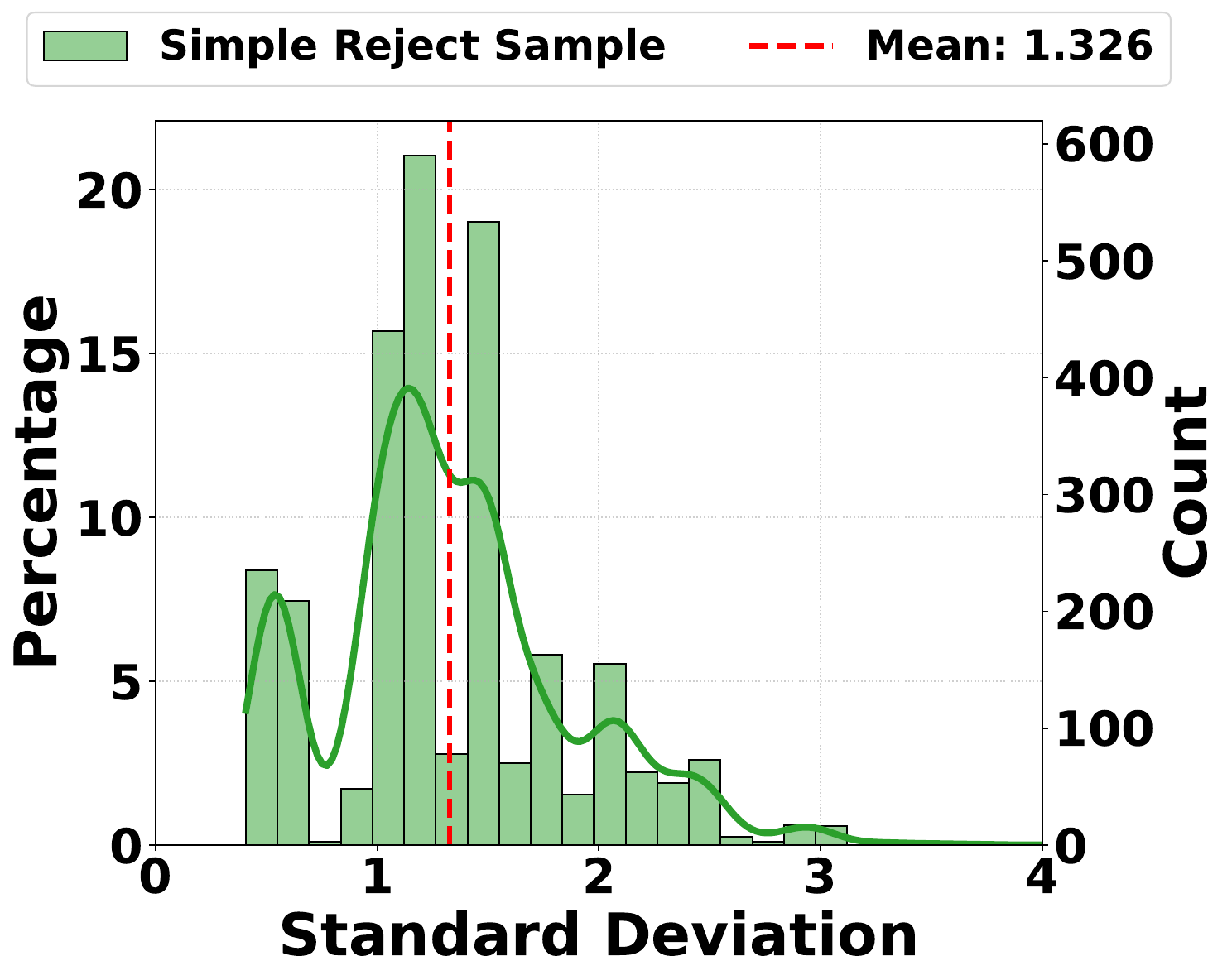}
    \end{subfigure}
    \hfill
    \begin{subfigure}[b]{0.19\textwidth}
        \centering
        \includegraphics[width=\textwidth]{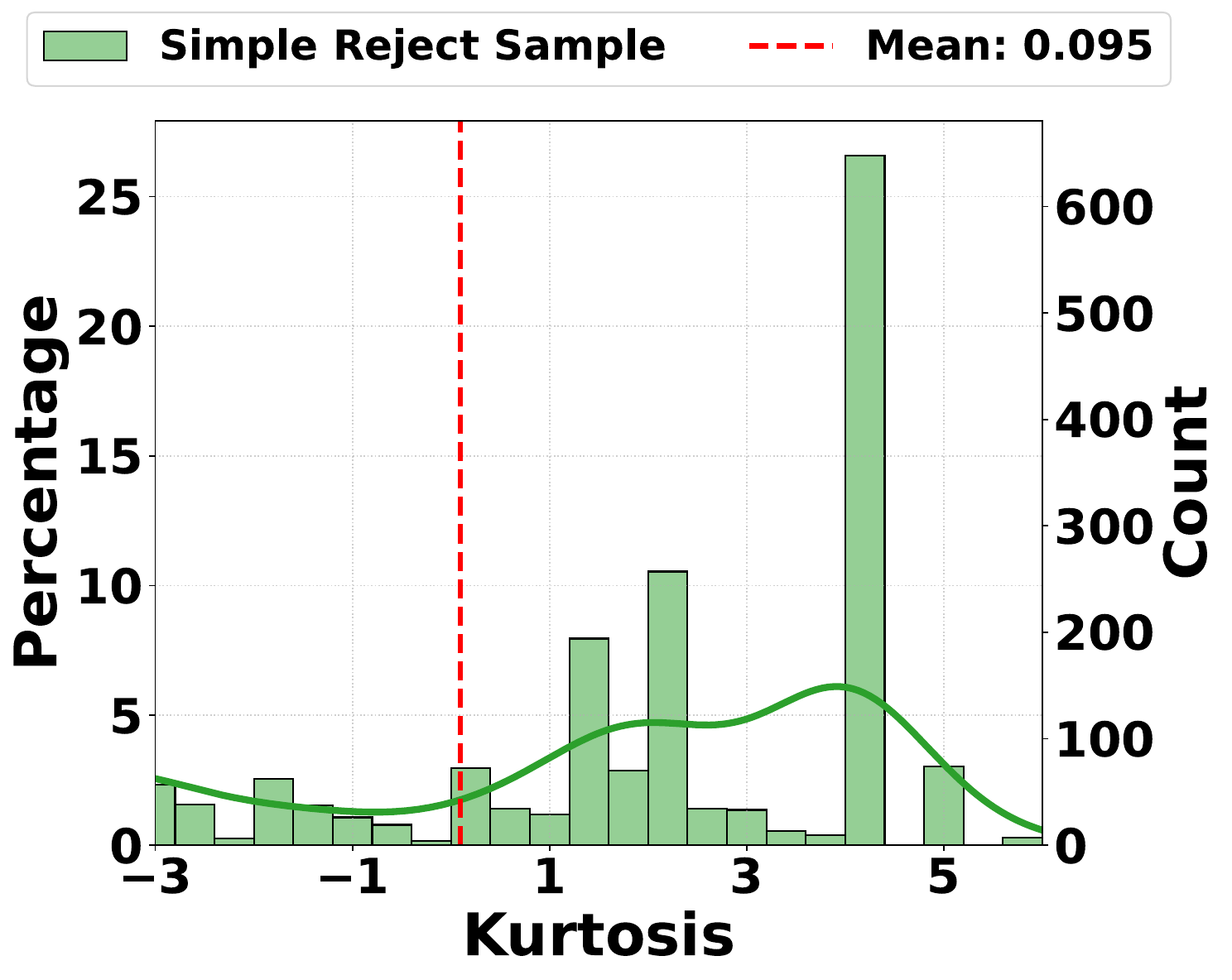}
    \end{subfigure}
    \hfill
    \begin{subfigure}[b]{0.19\textwidth}
        \centering
        \includegraphics[width=\textwidth]{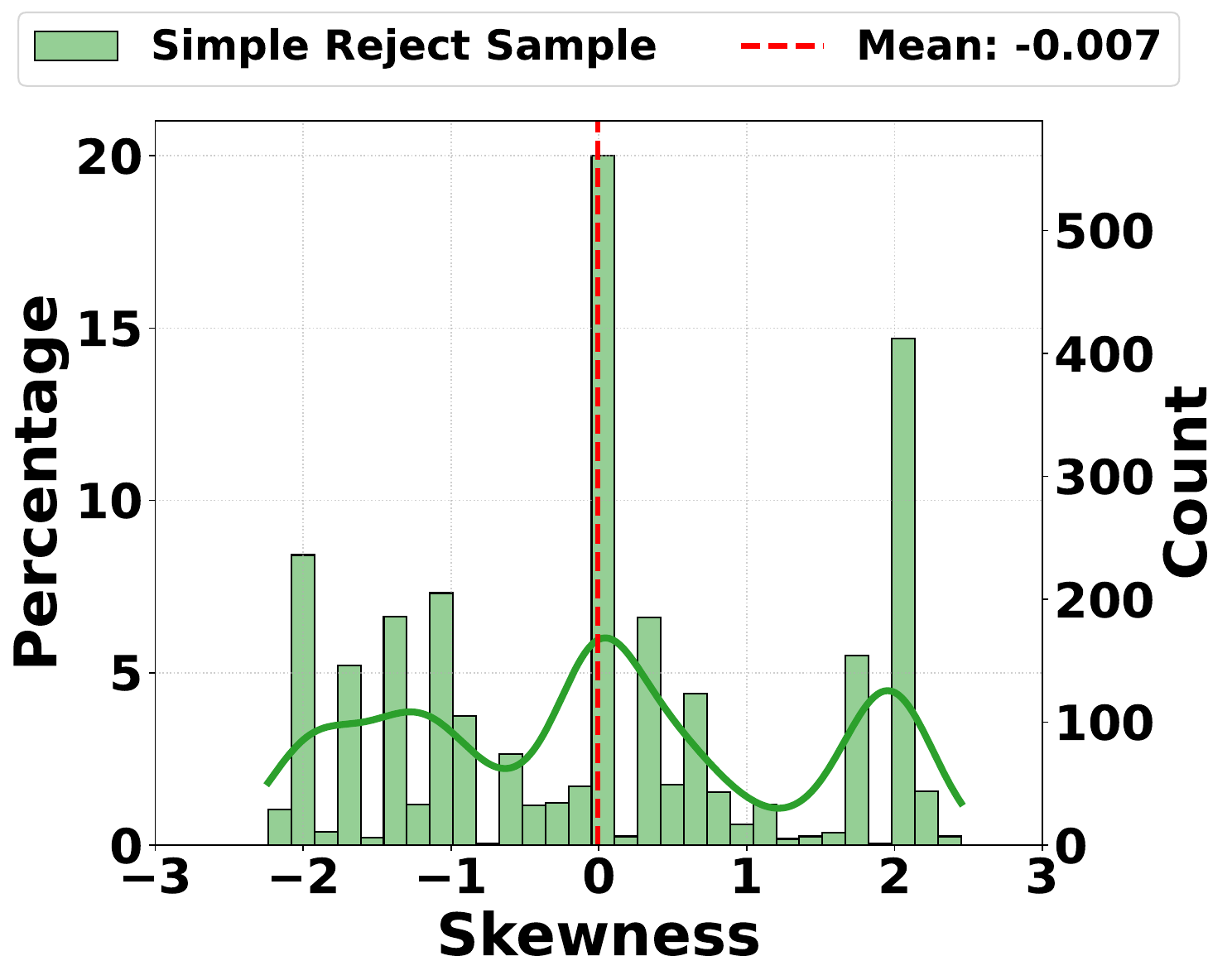}
    \end{subfigure}
    
    \caption{Statistical profile of review ratings across three sample categories (Hard Samples, Simple Accept, Simple Reject). Columns from left to right: Rating, Confidence, Standard Deviation, Kurtosis, Skewness.}
    \label{fig:rating_deiation}
\end{figure*}
%%%%%%%%%%%%%%%%%%%%%%%%%%%%%%%%

\section{Hard Sample Analysis}
\label{sec:hard_sample}

The significant performance gap between rating-based and text-based models  implies that numerical scores contain a deterministic signal for rejection that is effectively masked in textual reviews. To identify this missing signal, we analyze the statistical distribution of the Hard Samples (submissions rejected by the decision chair but predicted as Accept by text-based models) compared to easily classified samples. Figure \ref{fig:rating_deiation} presents a comprehensive statistical profile across five  metrics.

First, we examine the {Mean Rating} ($\mu$) and {Prediction Confidence} to understand the model's confusion. As illustrated in Figure \ref{fig:rating_deiation}, Simple Accept ($\mu \approx 6.77$) and Simple Reject ($\mu \approx 4.41$) samples exhibit distinct separability. In contrast, Hard Samples converge narrowly around $\mu=5.63$, positioning them precisely at the decision boundary. This ambiguity directly impacts the model's reliability: while simple samples elicit sharp, high-confidence predictions, Hard Samples suffer from a significant degradation in confidence ($\mu_{conf}=0.668$). This indicates that based on the average score alone, these submissions are indeed indistinguishable, forcing the model into a state of uncertainty.

However, the average score conceals the underlying structural conflict. To decouple legitimate disagreement from the specific rejection signal, we analyze the higher-order statistics:

\begin{itemize}
    \item {Standard Deviation:} Surprisingly, variance is not the distinguishing factor. Hard Samples ($\sigma=1.337$) share nearly identical divergence levels with Simple Rejects ($\sigma=1.326$). This suggests that high disagreement is a common trait of all rejected papers, not a unique signature of the hard cases.
    
    \item {Skewness:} This is the defining characteristic. Hard Samples display significant \textit{negative skewness} ($\mu=-0.486$), whereas Simple Rejects tend to have positive or near-zero skew. In the context of peer review, negative skewness implies a \textit{Veto Effect}: the majority of reviewers give high scores (pulling the distribution to the right), but a minority give very low scores (creating a long tail to the left).
    
    \item {Kurtosis:} We observe elevated kurtosis ($0.351$) in Hard Samples compared to the flatter distributions of simple cases. High kurtosis indicates that the low scores are not merely random noise but distinct outliers—representing strong, sharp objections.
\end{itemize}

In summary, Hard Samples are not simply mediocre papers with average scores. Instead, they are controversial submissions characterized by a {structural contradiction}: a generally positive consensus punctuated by a specific, fatal objection (High Kurtosis + Negative Skew). The rating-based model correctly identifies the decisive weight of the low score (the veto). In contrast, as we will discuss in the next section, text-based models are likely misled by the high volume of positive comments from the majority, failing to detect the subtle but lethal critique hidden in the politeness of the outlier.

%%%%%%%%%%%%%%%%%%%%%%%%%%%%%
\begin{table*}[!ht]
\centering
\renewcommand{\arraystretch}{1.3} 

\caption{Category and Aspect of text-based analysis.}
\label{tab:statistic_info_revised}

\resizebox{0.95\linewidth}{!}{%
\begin{tabular}{l p{12cm}} 
\toprule
\textbf{Category} & \textbf{Aspects} \\
\midrule
Evaluation & Result presentation, Result interpretation, Evaluation metrics, Ablation study, Baseline comparison, Figure table quality, Comparison fairness, Experimental thoroughness, Dataset appropriateness \\
\addlinespace % 增加行间距

Innovation \& Contribution & Technical soundness, Theoretical contribution, Practical significance, Novelty originality \\
\addlinespace

Methodology \& Technique & Methodology choice, Data quality, Theoretical foundation, Experimental design, Statistical analysis, Reproducibility \\
\addlinespace

Structure \& Logic & Contribution statement, Abstract completeness, Conclusion quality, Introduction background, Problem motivation, Literature review, Logical flow, Title quality \\
\addlinespace

Technical Details & Computational complexity, Implementation details, Scalability \\
\addlinespace

Writing \& Presentation & Writing quality, Citation reference, Technical language \\
\bottomrule
\end{tabular}
}
\end{table*}
%%%%%%%%%%%%%%%%%%%%%%%%%%%%%

\section{Sentiment Analysis of Review Comments}
\label{sec:sentiment_analysis}

To diagnose the failure of text-based models in capturing rejection signals, we conducted a fine-grained aspect-based sentiment analysis. We employed a hierarchical taxonomy of 6 macro and 33 sub-categories (Table \ref{tab:statistic_info_revised}) and quantified sentiment via a dependency-parsing augmented lexicon, aiming to decouple textual polarity from numerical ratings.

%%%%%%%%%%%%%%%%%%%%%%%%%%%%%%%%%%%
\begin{figure}[!t]
    \centering
    \includegraphics[width=1\linewidth]{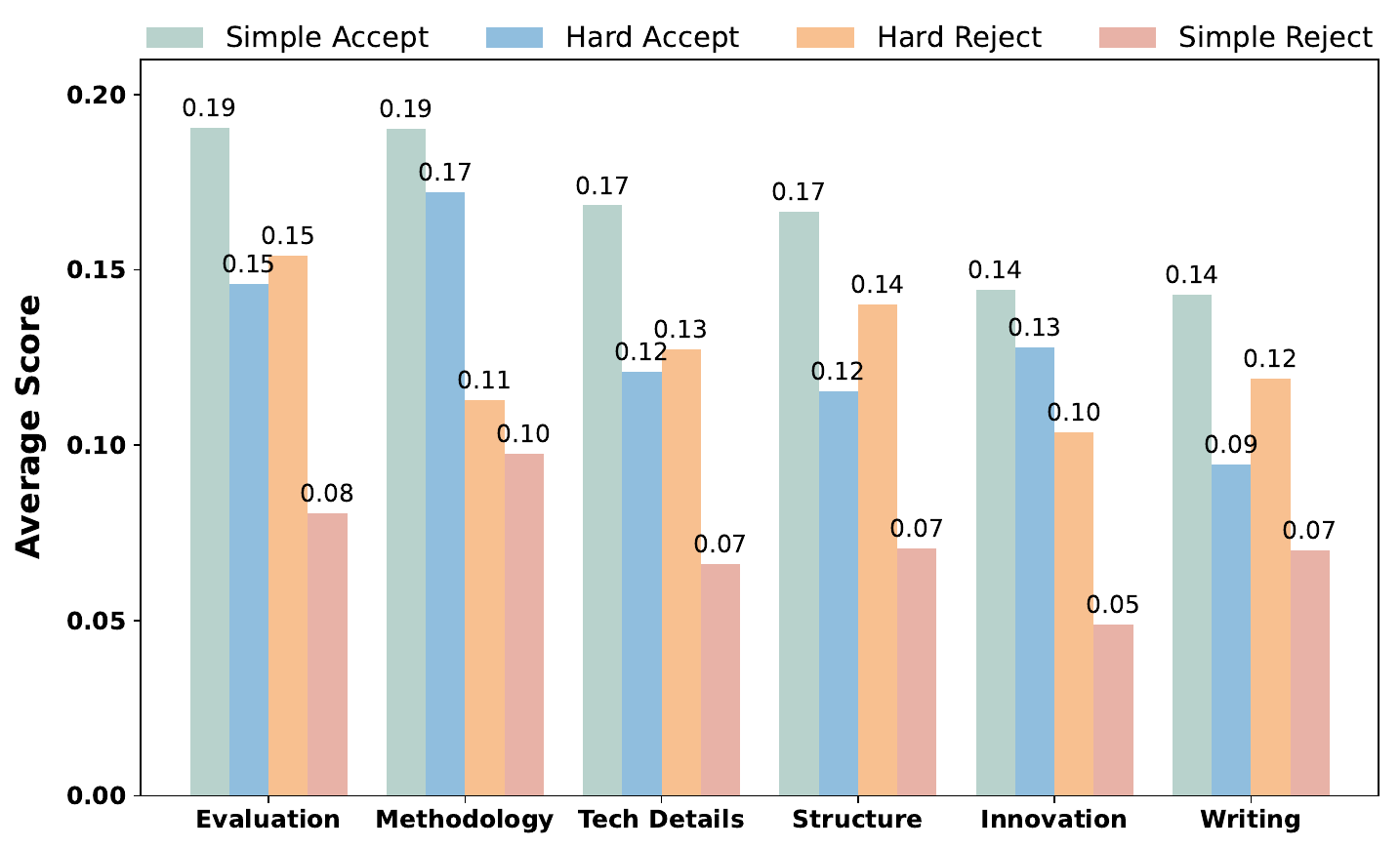}
    \caption{Average sentiment score for six high-level evaluation categories. Note that even for rejected papers (Hard Reject and Simple Reject), the average sentiment scores remain consistently positive ($>0$).}
    \label{fig:avg_score_by_category}
\end{figure}
%%%%%%%%%%%%%%%%%%%%%%%%%%%%%%%%%%%

Our analysis reveals a striking contrast between decision outcomes and textual sentiment.
First, as illustrated in Figure \ref{fig:avg_score_by_category}, the sentiment profiles of contradictory decisions exhibit a high degree of overlap. Specifically, the sentiment scores of Hard Samples align more closely with Simple Accepts ($\Delta=0.0418$) than with Simple Rejects ($\Delta=0.0558$). Crucially, despite the rejection decision, the average sentiment scores for Hard Rejects remain positive across all six macro-categories.
Second, this pervasive positivity is further corroborated at the fine-grained level. As shown in Figure \ref{fig:polarity_rates}, regardless of the final decision, positive comments dominate nearly all evaluation aspects ($>60\%$). Even in reviews for rejected submissions, the frequency of positive sentiment words consistently outweighs negative ones. In summary, the textual feedback for rejected papers structurally mimics the semantic patterns of accepted papers.

We attribute this ambiguity to the {Politeness Principle}. Unlike decisive numerical ratings, textual reviews are modulated by social pragmatics, where reviewers often soften criticism with praise. This ``polite masking'' creates a signal decoupling: high semantic positivity obscures the rejection ground truth. Consequently, text-based models are misled by this compliment noise, failing to detect fatal critiques and resulting in the observed performance gap.

%%%%%%%%%%%%%%%%%%%%%%%%%%%%%%%%%%%
\begin{figure}[!t]
    \centering
    \includegraphics[width=1\linewidth]{ratio_comparison_4_categories.pdf}
    \caption{Comparison of sentiments ratios across each aspect,  illustrating that positive sentiment dominates nearly all aspects, even for rejected papers (Hard/Simple Reject).}
    \label{fig:polarity_rates}
\end{figure}
%%%%%%%%%%%%%%%%%%%%%%%%%%%%%%%%%%%

\section{Conclusion}
In this work, we constructed a comprehensive ICLR 2021-2025 dataset to quantify the predictability of  review decisions. Our benchmarking reveals a significant {Signal Decoupling}: numerical scores achieve 91\% accuracy, whereas text-based models stagnate at 81\%.
Our diagnostic analysis attributes this gap to a structural contradiction between the two modalities. Statistically, the Hard Samples are defined by {negative skewness} and high kurtosis, representing a veto scenario where a specific objection outweighs the majority consensus. Semantically, this decisive rejection signal is obscured by the {Politeness Principle}. We observe that reviewers habitually cloak critical negative feedback within predominantly positive language, creating a semantic distribution that mimics accepted papers.
This study demonstrates that while LLMs excel at understanding content, they struggle to decode the social nuance of polite rejection. Future research must move beyond semantic sentiment to capture the implicit pragmatic signals that drive human decision-making.

\bibliographystyle{acl_natbib}
\bibliography{reference}

\begin{thebibliography}{24}
\providecommand{\natexlab}[1]{#1}

\bibitem[{Beltagy et~al.(2019)Beltagy, Lo, and Cohan}]{beltagy-etal-2019-scibert}
Iz~Beltagy, Kyle Lo, and Arman Cohan. 2019.
\newblock \href {https://doi.org/10.18653/v1/D19-1371} {{S}ci{BERT}: A pretrained language model for scientific text}.
\newblock In \emph{Proceedings of the 2019 Conference on Empirical Methods in Natural Language Processing and the 9th International Joint Conference on Natural Language Processing (EMNLP-IJCNLP)}, pages 3615--3620, Hong Kong, China. Association for Computational Linguistics.

\bibitem[{Bharti et~al.(2023)Bharti, Navlakha, Agarwal, and Ekbal}]{10.1007/s10579-023-09662-3}
Prabhat~Kumar Bharti, Meith Navlakha, Mayank Agarwal, and Asif Ekbal. 2023.
\newblock \href {https://doi.org/10.1007/s10579-023-09662-3} {Politepeer: does peer review hurt? a dataset to gauge politeness intensity in the peer reviews}.
\newblock 58(4):1291–1313.

\bibitem[{Brown(1987)}]{brown1987politeness}
Penelope Brown. 1987.
\newblock \emph{Politeness: Some universals in language usage}, volume~4.
\newblock Cambridge university press.

\bibitem[{Chakraborty et~al.(2020)Chakraborty, Goyal, and Mukherjee}]{10.1145/3383583.3398541}
Souvic Chakraborty, Pawan Goyal, and Animesh Mukherjee. 2020.
\newblock \href {https://doi.org/10.1145/3383583.3398541} {Aspect-based sentiment analysis of scientific reviews}.
\newblock In \emph{Proceedings of the ACM/IEEE Joint Conference on Digital Libraries in 2020}, JCDL '20, page 207–216, New York, NY, USA. Association for Computing Machinery.

\bibitem[{Deng et~al.(2020)Deng, Peng, Xia, Li, He, and Yu}]{deng-etal-2020-hierarchical}
Zhongfen Deng, Hao Peng, Congying Xia, Jianxin Li, Lifang He, and Philip Yu. 2020.
\newblock \href {https://doi.org/10.18653/v1/2020.coling-main.555} {Hierarchical bi-directional self-attention networks for paper review rating recommendation}.
\newblock In \emph{Proceedings of the 28th International Conference on Computational Linguistics}, pages 6302--6314, Barcelona, Spain (Online). International Committee on Computational Linguistics.

\bibitem[{Fernandes and Vaz-de Melo(2022)}]{10.1145/3529372.3530935}
Gustavo~L\'{u}cius Fernandes and Pedro O.~S. Vaz-de Melo. 2022.
\newblock \href {https://doi.org/10.1145/3529372.3530935} {Between acceptance and rejection: challenges for an automatic peer review process}.
\newblock New York, NY, USA. Association for Computing Machinery.

\bibitem[{Fernandes and Vaz-de Melo(2024)}]{fernandes2024enhancing}
Gustavo~L{\'u}cius Fernandes and Pedro~OS Vaz-de Melo. 2024.
\newblock Enhancing the examination of obstacles in an automated peer review system.
\newblock \emph{International Journal on Digital Libraries}, 25(2):341--364.

\bibitem[{Gao et~al.(2019)Gao, Eger, Kuznetsov, Gurevych, and Miyao}]{gao-etal-2019-rebuttal}
Yang Gao, Steffen Eger, Ilia Kuznetsov, Iryna Gurevych, and Yusuke Miyao. 2019.
\newblock \href {https://doi.org/10.18653/v1/N19-1129} {Does my rebuttal matter? insights from a major {NLP} conference}.
\newblock In \emph{Proceedings of the 2019 Conference of the North {A}merican Chapter of the Association for Computational Linguistics: Human Language Technologies, Volume 1 (Long and Short Papers)}, pages 1274--1290, Minneapolis, Minnesota. Association for Computational Linguistics.

\bibitem[{Ghosal et~al.(2019)Ghosal, Verma, Ekbal, and Bhattacharyya}]{ghosal-etal-2019-deepsentipeer}
Tirthankar Ghosal, Rajeev Verma, Asif Ekbal, and Pushpak Bhattacharyya. 2019.
\newblock \href {https://doi.org/10.18653/v1/P19-1106} {{D}eep{S}enti{P}eer: Harnessing sentiment in review texts to recommend peer review decisions}.
\newblock In \emph{Proceedings of the 57th Annual Meeting of the Association for Computational Linguistics}, pages 1120--1130, Florence, Italy. Association for Computational Linguistics.

\bibitem[{Hua et~al.(2019)Hua, Hu, and Wang}]{hua-etal-2019-argument-generation}
Xinyu Hua, Zhe Hu, and Lu~Wang. 2019.
\newblock \href {https://doi.org/10.18653/v1/P19-1255} {Argument generation with retrieval, planning, and realization}.
\newblock In \emph{Proceedings of the 57th Annual Meeting of the Association for Computational Linguistics}, pages 2661--2672, Florence, Italy. Association for Computational Linguistics.

\bibitem[{Huang et~al.(2023)Huang, bin Huang, Bu, Cao, Shen, and Cheng}]{HUANG2023101427}
Junjie Huang, Win bin Huang, Yi~Bu, Qi~Cao, Huawei Shen, and Xueqi Cheng. 2023.
\newblock \href {https://doi.org/10.1016/j.joi.2023.101427} {What makes a successful rebuttal in computer science conferences?: A perspective on social interaction}.
\newblock \emph{Journal of Informetrics}, 17(3):101427.

\bibitem[{Idahl and Ahmadi(2025)}]{idahl-ahmadi-2025-openreviewer}
Maximilian Idahl and Zahra Ahmadi. 2025.
\newblock \href {https://doi.org/10.18653/v1/2025.naacl-demo.44} {{O}pen{R}eviewer: A specialized large language model for generating critical scientific paper reviews}.
\newblock In \emph{Proceedings of the 2025 Conference of the Nations of the Americas Chapter of the Association for Computational Linguistics: Human Language Technologies (System Demonstrations)}, pages 550--562, Albuquerque, New Mexico. Association for Computational Linguistics.

\bibitem[{Kang et~al.(2018)Kang, Ammar, Dalvi, van Zuylen, Kohlmeier, Hovy, and Schwartz}]{kang-etal-2018-dataset}
Dongyeop Kang, Waleed Ammar, Bhavana Dalvi, Madeleine van Zuylen, Sebastian Kohlmeier, Eduard Hovy, and Roy Schwartz. 2018.
\newblock \href {https://doi.org/10.18653/v1/N18-1149} {A dataset of peer reviews ({P}eer{R}ead): Collection, insights and {NLP} applications}.
\newblock In \emph{Proceedings of the 2018 Conference of the North {A}merican Chapter of the Association for Computational Linguistics: Human Language Technologies, Volume 1 (Long Papers)}, pages 1647--1661, New Orleans, Louisiana. Association for Computational Linguistics.

\bibitem[{Kargaran et~al.(2025)Kargaran, Nikeghbal, Yang, and Ousidhoum}]{kargaran2025insights}
Amir~Hossein Kargaran, Nafiseh Nikeghbal, Jing Yang, and Nedjma Ousidhoum. 2025.
\newblock Insights from the iclr peer review and rebuttal process.
\newblock \emph{arXiv preprint arXiv:2511.15462}.

\bibitem[{Leng et~al.(2019)Leng, Yu, and Xiong}]{10.1145/3340555.3353766}
Youfang Leng, Li~Yu, and Jie Xiong. 2019.
\newblock \href {https://doi.org/10.1145/3340555.3353766} {Deepreviewer: Collaborative grammar and innovation neural network for automatic paper review}.
\newblock In \emph{2019 International Conference on Multimodal Interaction}, ICMI '19, page 395–403, New York, NY, USA. Association for Computing Machinery.

\bibitem[{Li et~al.(2025)Li, Gu, Kung, Xia, Kong, Sui, Peng et~al.}]{li2025llm}
Rui Li, Jia-Chen Gu, Po-Nien Kung, Heming Xia, Xiangwen Kong, Zhifang Sui, Nanyun Peng, and 1 others. 2025.
\newblock Llm-reval: Can we trust llm reviewers yet?
\newblock \emph{arXiv preprint arXiv:2510.12367}.

\bibitem[{Lu et~al.(2025)Lu, Xu, Li, Ding, and Meng}]{lu2025agent}
Kai Lu, Shixiong Xu, Jinqiu Li, Kun Ding, and Gaofeng Meng. 2025.
\newblock \href {https://openreview.net/forum?id=s7HUJamWqX} {Agent reviewers: Domain-specific multimodal agents with shared memory for paper review}.
\newblock In \emph{Forty-second International Conference on Machine Learning}.

\bibitem[{Meng et~al.(2023)Meng, Han, Zhong, Zhou, and Zhang}]{MinghuiMeng2023AspectbasedSA}
Minghui Meng, Ruxue Han, Jiangtao Zhong, Haomin Zhou, and Chengzhi Zhang. 2023.
\newblock \href {https://doi.org/10.59494/dsi.2023.1.4} {Aspect-based sentiment analysis of online peer reviews and prediction of paper acceptance results}.
\newblock \emph{Data Science and Informetrics}, 81 1 1:0.

\bibitem[{Ribeiro et~al.(2021)Ribeiro, Sizo, Lopes~Cardoso, and Reis}]{10.1007/978-3-030-86230-5_60}
Ana~Carolina Ribeiro, Amanda Sizo, Henrique Lopes~Cardoso, and Lu\'{\i}s~Paulo Reis. 2021.
\newblock \href {https://doi.org/10.1007/978-3-030-86230-5_60} {Acceptance decision prediction in peer-review through sentiment analysis}.
\newblock In \emph{Progress in Artificial Intelligence: 20th EPIA Conference on Artificial Intelligence, EPIA 2021, Virtual Event, September 7–9, 2021, Proceedings}, page 766–777, Berlin, Heidelberg. Springer-Verlag.

\bibitem[{Wang and Wan(2018)}]{10.1145/3209978.3210056}
Ke~Wang and Xiaojun Wan. 2018.
\newblock \href {https://doi.org/10.1145/3209978.3210056} {Sentiment analysis of peer review texts for scholarly papers}.
\newblock In \emph{The 41st International ACM SIGIR Conference on Research \& Development in Information Retrieval}, SIGIR '18, page 175–184, New York, NY, USA. Association for Computing Machinery.

\bibitem[{Wang et~al.(2020)Wang, Zeng, Huang, Knight, Ji, and Rajani}]{wang-etal-2020-reviewrobot}
Qingyun Wang, Qi~Zeng, Lifu Huang, Kevin Knight, Heng Ji, and Nazneen~Fatema Rajani. 2020.
\newblock \href {https://doi.org/10.18653/v1/2020.inlg-1.44} {{R}eview{R}obot: Explainable paper review generation based on knowledge synthesis}.
\newblock In \emph{Proceedings of the 13th International Conference on Natural Language Generation}, pages 384--397, Dublin, Ireland. Association for Computational Linguistics.

\bibitem[{Zhou et~al.(2024)Zhou, Chen, and Yu}]{zhou-etal-2024-llm}
Ruiyang Zhou, Lu~Chen, and Kai Yu. 2024.
\newblock \href {https://aclanthology.org/2024.lrec-main.816/} {Is {LLM} a reliable reviewer? a comprehensive evaluation of {LLM} on automatic paper reviewing tasks}.
\newblock In \emph{Proceedings of the 2024 Joint International Conference on Computational Linguistics, Language Resources and Evaluation (LREC-COLING 2024)}, pages 9340--9351, Torino, Italia. ELRA and ICCL.

\bibitem[{Zhu et~al.(2025)Zhu, Xiong, Ma, Lu, Liu, and Li}]{Zhu2025WhenYR}
Changjia Zhu, Junjie Xiong, Renkai Ma, Zhicong Lu, Yao Liu, and Lingyao Li. 2025.
\newblock \href {https://api.semanticscholar.org/CorpusID:281310346} {When your reviewer is an llm: Biases, divergence, and prompt injection risks in peer review}.
\newblock \emph{ArXiv}, abs/2509.09912.

\bibitem[{Zhuang et~al.(2025)Zhuang, Chen, Xu, Jiang, and Lin}]{ZHUANG2025103332}
Zhenzhen Zhuang, Jiandong Chen, Hongfeng Xu, Yuwen Jiang, and Jialiang Lin. 2025.
\newblock \href {https://doi.org/10.1016/j.inffus.2025.103332} {Large language models for automated scholarly paper review: A survey}.
\newblock \emph{Information Fusion}, 124:103332.

\end{thebibliography}
\end{document}